\documentclass[10pt,twocolumn,letterpaper]{article}

\usepackage{cvpr}              

\usepackage{graphicx}
\usepackage{amsmath}
\usepackage{amssymb}
\usepackage{booktabs}
\usepackage{bm}
\usepackage{diagbox}
\usepackage{graphbox} 
\usepackage{color, colortbl}

\definecolor{Gray}{gray}{0.9}

\DeclareMathOperator{\arcosh}{arcosh}

\usepackage[pagebackref,breaklinks,colorlinks]{hyperref}

\usepackage[capitalize]{cleveref}
\crefname{section}{Sec.}{Secs.}
\Crefname{section}{Section}{Sections}
\Crefname{table}{Table}{Tables}
\crefname{table}{Tab.}{Tabs.}

\def\cvprPaperID{9445} 
\def\confName{CVPR}
\def\confYear{2022}

\newcommand{\tb}[3]{\setlength{\tabcolsep}{#2mm}\begin{tabular}{#1}#3\end{tabular}}

\begin{document}

\title{CO-SNE: Dimensionality Reduction and Visualization for Hyperbolic Data}

\author{
\tb{@{}ccc@{}}{12}{
Yunhui Guo&  
Haoran Guo& 
Stella X. Yu\\
}\\
UC Berkeley / ICSI
}

\maketitle

\begin{abstract}
Hyperbolic space can naturally embed hierarchies that often exist in real-world data and semantics. While high-dimensional hyperbolic embeddings lead to better representations, most hyperbolic models utilize low-dimensional embeddings, due to non-trivial optimization and visualization of high-dimensional hyperbolic data. 

We propose CO-SNE, which extends the Euclidean space visualization tool, t-SNE, to hyperbolic space. Like t-SNE, it converts distances between data points to joint probabilities and tries to minimize the Kullback-Leibler divergence between the joint probabilities of high-dimensional data $X$ and low-dimensional embedding $Y$. However, unlike Euclidean space, hyperbolic space is inhomogeneous: A volume could contain a lot more points at a location far from the origin. CO-SNE thus uses hyperbolic normal distributions for $X$ and hyperbolic \underline{C}auchy instead of t-SNE's Student's t-distribution for $Y$, and it additionally seeks to preserve $X$'s individual distances to the \underline{O}rigin in $Y$.

We apply CO-SNE to naturally hyperbolic data and supervisedly learned hyperbolic features. Our results demonstrate that CO-SNE deflates high-dimensional hyperbolic data into a low-dimensional space without losing their hyperbolic characteristics, significantly outperforming popular visualization tools such as PCA, t-SNE, UMAP, and HoroPCA which is also designed for hyperbolic data.
\end{abstract}

\section{Introduction}
\label{sec:intro}
\begin{figure}
    \centering
    \includegraphics[width=0.46\textwidth]{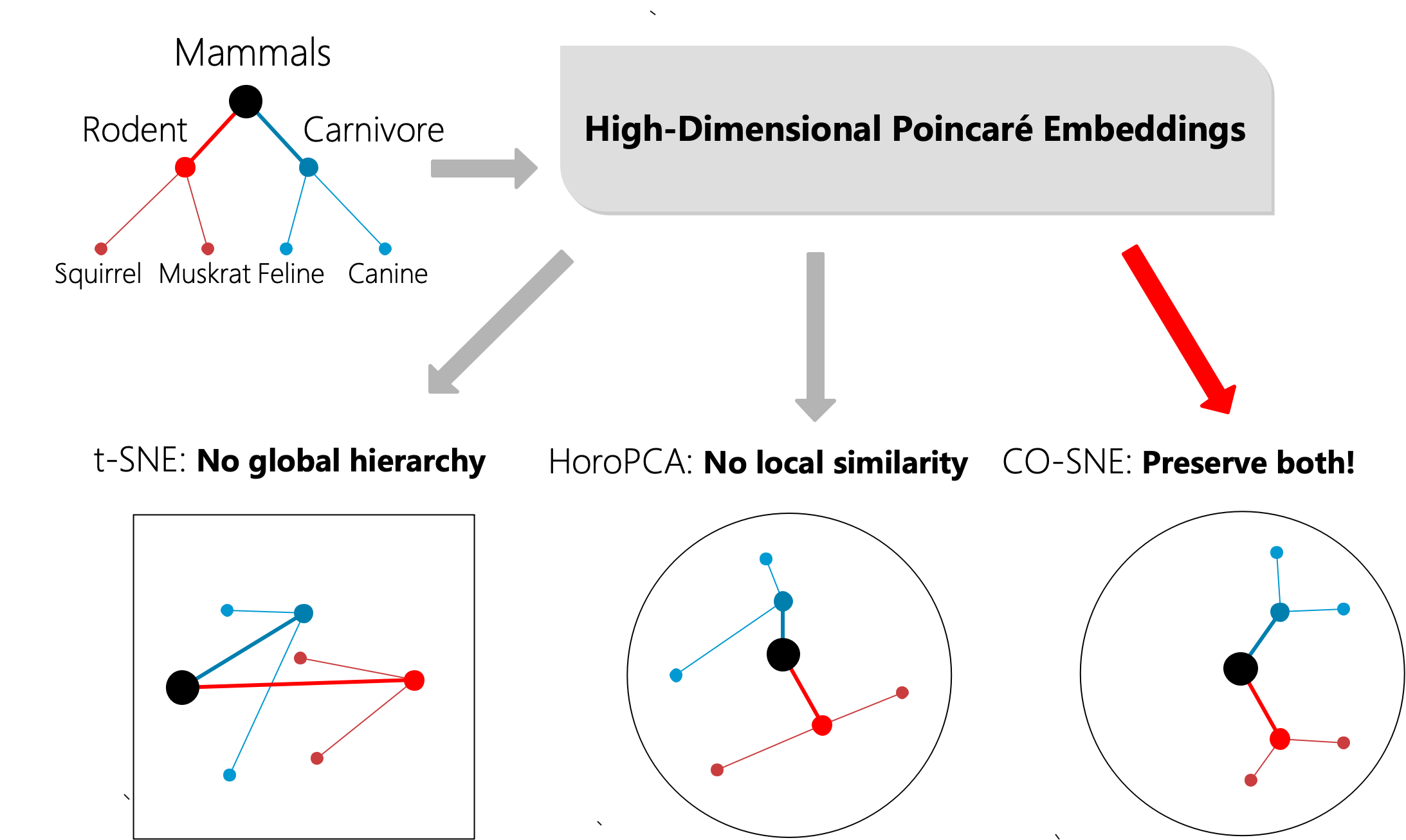}
    \caption{Our CO-SNE method deflates a high-dimensional hyperbolic representation while preserving their global hierarchy and local similarities. We generate the Poincaré embeddings \cite{nickel2017poincar} of the $mammal$ subtree from WordNet in a five-dimensional hyperbolic space. We apply the standard t-SNE \cite{van2008visualizing}, the recently proposed HoroPCA \cite{chami2021horopca} and our proposed CO-SNE method to visualize the embeddings in a two-dimensional Euclidean (in t-SNE) or hyperbolic space (in HoroPCA and CO-SNE). CO-SNE preserves the global hierarchy (root node is in the center and leaf nodes are close to boundary) and the local similarity (sibling nodes are close) in the two-dimensional embeddings.}
    \label{fig:teaser}
\end{figure}

Datasets with hierarchical structures are ubiquitous. Social networks \cite{gupte2011finding} and complex networks \cite{alanis2016efficient} are representative examples of hierarchical data. Euclidean space cannot embed entities in such hierarchical datasets without distortion. Hyperbolic space, a non-Euclidean space with constant negative curvature, has been widely used for embedding hierarchical data since hyperbolic metric can closely approximate the tree metric. Hyperbolic space has thus been used in the representation learning of word embeddings \cite{nickel2017poincar} (Figure \ref{fig:teaser}) and visual inputs \cite{khrulkov2020hyperbolic,mathieu2019continuous}. Several algorithms also directly operate on hyperbolic space  \cite{ganea2018hyperbolic,cho2019large,weber2020robust}. 

Higher embedding dimension can generally lead to better hyperbolic representation quality \cite{nickel2017poincar,hsu2020learning}. However, learning with two-dimensional hyperbolic space is prevalent \cite{nickel2017poincar,hsu2020learning}. One of the reasons is the ease of visualization. There are several isometrically equivalent models for representing hyperbolic space. The Poincar\'e ball model is arguably the most widely used model in hyperbolic representation learning \cite{nickel2017poincar,hsu2020learning,guo2021free}. With the Poincar\'e ball model, we can easily visualize two-dimensional hyperbolic embeddings within a unit Euclidean circle. However, to visualize high-dimensional hyperbolic data is not easy, as most visualization methods assume the data exist in Euclidean space.

Embeddings in Poincar\'e ball have two notable properties: {\bf 1)} The embeddings have a global hierarchical structure. Root nodes are in the center of the ball and leaf nodes are close to the boundary of the ball. {\bf 2)} The embeddings possess a local similarity structure. Sibling nodes should be close in the embedding space.

t-SNE \cite{van2008visualizing} is a popular visualization tool for visualizing high-dimensional Euclidean data. However, t-SNE cannot preserve the global hierarchy of the hyperbolic embeddings. HoroPCA \cite{chami2021horopca} is recently proposed as an extension of PCA on hyperbolic space. However, HoroPCA cannot preserve the local similarity of the hyperbolic embeddings. In this paper, we propose CO-SNE which can preserve both the global hierarchy and local similarity of high-dimensional hyperbolic embeddings in a low- dimensional hyperbolic space (\ref{fig:teaser}). In Figure \ref{fig:teaser}, the $mammal$ subtree from WordNet \cite{miller1995wordnet} is embedded in a five-dimensional hyperbolic space via Poincar\'e embeddings \cite{nickel2017poincar}. We use t-SNE \cite{van2008visualizing}, HoroPCA \cite{chami2021horopca} and CO-SNE to visualize the embeddings in a two-dimensional space. CO-SNE can preserve the structure of the data well.

In CO-SNE, for maintaining local similarity structure, we adopt the same idea as in the standard t-SNE to minimize the KL-divergence between the high-dimensional similarities and the low-dimensional similarities. We adopt hyperbolic versions of the normal distribution and \underline{C}auchy distribution to compute the similarities. To maintain the global hierarchical structure, we adopt a distance loss function which seeks to preserve the individual distances of high-dimensional hyperbolic embeddings to the \underline{O}rigin in low-dimensional hyperbolic space. 

In summary, we make the following contributions,

\begin{itemize}
    \item We propose CO-SNE which can represent high-dimensional hyperbolic datapoints in a low-dimensional hyperbolic space while maintaining the local similarity and the global hierarchical structure. 
    
    \item We propose to use hyperbolic Cauchy distribution for computing low-dimensional similarities which is crucial for producing good visualization in hyperbolic space.
    
    \item We apply CO-SNE to visualize synthetic hyperbolic data, hierarchical biological datasets and hyperbolic embeddings learned by supervised and unsupervised learning methods to better understand high-dimensional hyperbolic data. Across all the cases, CO-SNE produces much better visualization than the baselines.
\end{itemize}

\section{Related Work}

\indent \textbf{Large-margin Classification} \cite{cho2019large} in hyperbolic space was proposed by changing the distance function from Euclidean to hyperbolic. Robust large-margin classification \cite{weber2020robust} in hyperbolic space is also proposed and has the first theoretical guarantees for learning a hyperbolic classifier.

\indent \textbf{Hyperbolic Neural Networks (HNNs)} \cite{ganea2018hyperbolic} rewrite multinomial logistic regression (MLR), fully connected layers and Recurrent Neural Networks for hyperbolic embeddings via gyrovector space operations \cite{ungar2008gyrovector}. In a follow-up work, Hyperbolic Neural Networks++ \cite{shimizu2020hyperbolic} introduces hyperbolic convolutional layers. Hyperbolic attention networks \cite{gulcehre2018hyperbolic} are proposed by extending attention operations to hyperbolic space in a manner similar to \cite{ganea2018hyperbolic}. Hyperbolic graph neural networks \cite{liu2019hyperbolic} are further proposed by altering the geometry of Graph Neural Networks (GNNs) \cite{zhou2020graph} to hyperbolic space. Hyperbolic neural networks have also been used for visual inputs and achieve better results than Euclidean neural networks on tasks such as few-shot classification and person re-identification \cite{khrulkov2020hyperbolic}.

\indent \textbf{Unsupervised Learning Methods in Hyperbolic Space} are also proposed recently. Poincar\'e embeddings are proposed to embed words and relations in hyperbolic space \cite{nickel2017poincar}.  \cite{nagano2019wrapped} proposed a generalized version of the normal distribution on hyperbolic space called wrapped normal distribution. The proposed wrapped normal distribution is used as the latent space for constructing hyperbolic variational autoencoders (VAEs) \cite{kingma2013auto}. A similar idea is also adopted in \cite{mathieu2019continuous} to construct Poincar\'e VAEs. Unsupervised 3D segmentation \cite{hsu2020learning} and instance segmentation \cite{weng2021unsupervised} are achieved in hyperbolic space via hierarchical triplet loss.

\indent \textbf{Data Visualization} is the process of generating low-dimensional representation of each high-dimensional datapoint. A good visualization should maintain the interesting structure of the data presented in high-dimensional space. t-distributed stochastic neighbor embedding (t-SNE) \cite{van2008visualizing} is arguably the most widely used tool for data visualization. t-SNE attempts to maintain the local similarities of the high-dimensional datapoints in the low-dimensional space. More recently, UMAP \cite{mcinnes2018umap} is proposed as a manifold learning technique for dimensionality reduction and visualization. Compared with t-SNE, UMAP can better preserve the global structure of the high-dimensional data. 

\indent \textbf{Dimensionality Reduction} methods also can be used for data visualization. Principal Component Analysis (PCA \cite{jolliffe2016principal} is a commonly used dimensionality reduction technique. PCA aims at maintaining the maximum amount of variation of information in the low-dimensional space. Isomap \cite{tenenbaum2000global} is a non-linear dimensionality reduction method which attempts to preserve local structures. LLE \cite{roweis2000nonlinear} is another popular non-linear dimensionality reduction method that can produce neighborhood preserving embeddings of high-dimensional data. 
    
Notably, no existing data visualization methods are designed for visualizing high-dimensional hyperbolic data. Our proposed CO-SNE method can be used to visualize high-dimensional hyperbolic embeddings.

\begin{figure}
    \centering
    \includegraphics[width=0.4\textwidth]{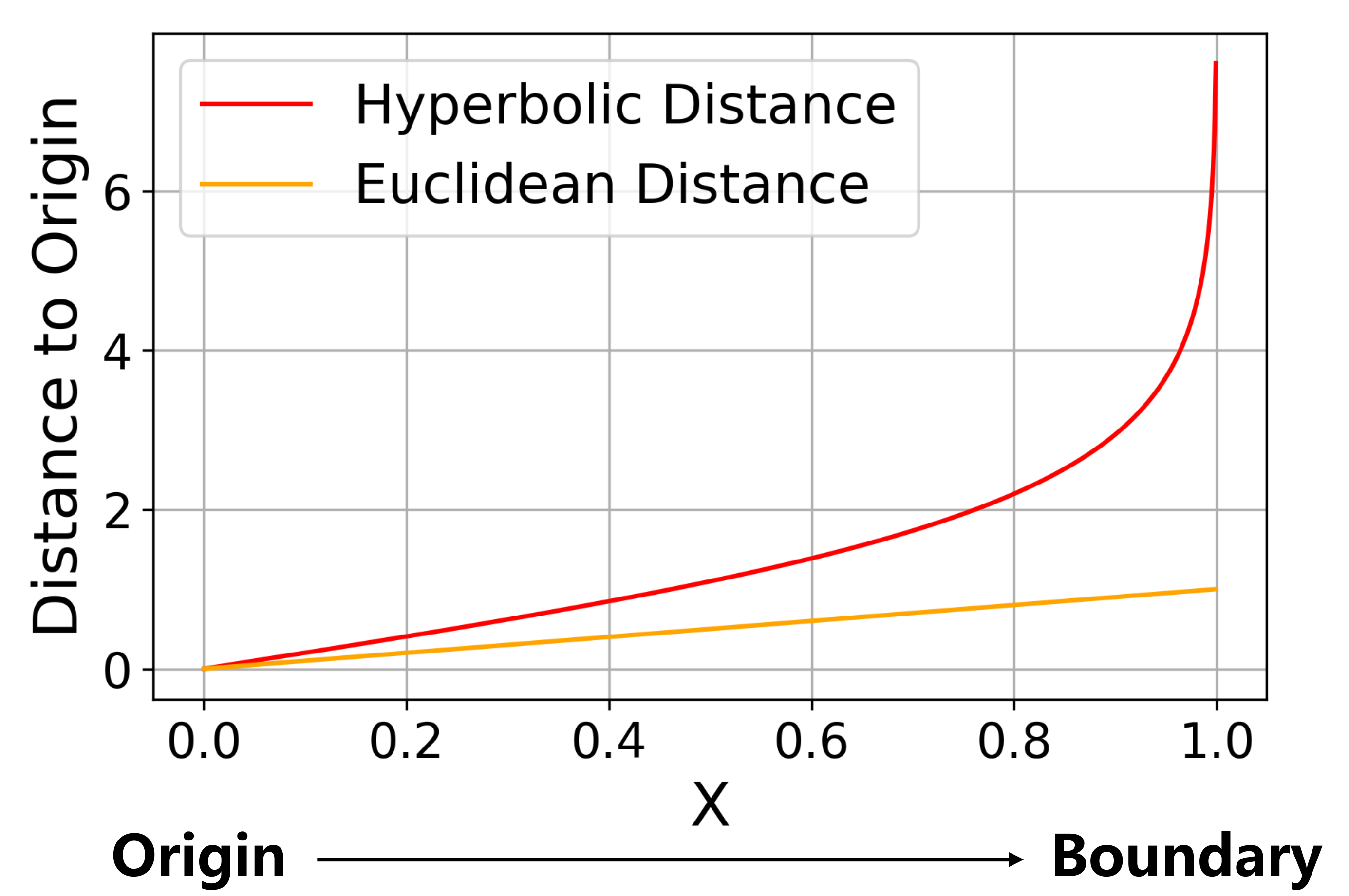}
    \caption{The comparison of Euclidean distance and hyperbolic distance as we move towards the boundary of the Poincar\'e ball.  }
    \label{fig:dist}
\end{figure}

\section{CO-SNE}
We present the proposed CO-SNE method which can faithfully represent high-dimensional hyperbolic data in a low-dimensional hyperbolic space. In Section \ref{sec:p_ball}, we review the basics of Poincar\'e ball model for hyperbolic space. In Section \ref{sec:t-sne}, we review t-distributed Stochastic neighbor embedding (t-SNE). In Section \ref{sec:normal} and \ref{sec:student_t}, we introduce the hyperbolic version of normal distribution and Student's t-distribution. In Section \ref{sec:not_heavy}, we discuss the issue of using hyperbolic Student's t-distribution for computing low-dimensional similarities and propose to use hyperbolic Cauchy distribution. In Section \ref{sec:rcl}, we present the distance loss for maintaining global hierarchy. In Section \ref{sec:criterion}, we present the optimization details of CO-SNE.

\subsection{Poincar\'e Ball Model for Hyperbolic Space}
\label{sec:p_ball}
 A hyperbolic space is a Riemannian manifold with a constant negative curvature. There are several isometrically equivalent models for representing hyperbolic space. The Poincar\'e ball model is the most commonly used one in hyperbolic representation learning \cite{nickel2017poincar,ganea2018hyperbolic}. The $n$-dimensional Poincar\'e  ball model is defined as $(\mathbb{B}^n, \mathfrak{g}_\mathbf{x})$, where $\mathbb{B}^n$ = $\{\mathbf{x} \in \mathbb{R}^n: \lVert \mathbf{x} \rVert < 1 \}$ and $\mathfrak{g}_\mathbf{x}  = (\gamma_\mathbf{x})^2 I_n $ is the Riemannian metric tensor. $\gamma_\mathbf{x} = \frac{2}{1- \lVert \mathbf{x} \rVert^2}$ is the conformal factor and $I_n$ is the Euclidean metric tensor. Given two points $\bm{u} \in \mathbb{B}^n$ and $\bm{v} \in \mathbb{B}^n$, the hyperbolic distance between them is defined as,
\begin{equation}d_{\mathbb{B}^n}(\bm{u}, \bm{v}) = \arcosh\left(1 + 2\frac{\lVert\bm{u}-\bm{v}\rVert^2}{(1-\lVert\bm{u}\rVert^2)(1-\lVert\bm{v}\rVert^2)}\right)\label{eq1}\end{equation}
where $\arcosh$ is the inverse hyperbolic cosine function and $\lVert \cdot \rVert$ is the usual Euclidean norm. Different from Euclidean distance, hyperbolic distance grows exponentially fast as we move the points towards the boundary of the Poincar\'e ball as in Figure \ref{fig:dist}.

\subsection{t-SNE}
\label{sec:t-sne}
t-SNE \cite{van2008visualizing} begins by mapping high-dimensional distances between datapoints to similarity values. The similarity values are either conditional or joint probabilities based on the probability a point will pick another as its neighbor if neighbors are picked in proportion to the probability density of a distribution centered at that point. t-SNE defines the conditional probability $p_{j|i}$, the probability that the point $\mathbf{x}_i$ will pick a point $\mathbf{x}_j$ as its neighbor, using a normal distribution centered at the point $\mathbf{x}_i$. t-SNE then defines the joint probability distribution $P$, by setting $p_{ij} = \frac{p_{i|j} + p_{j|i}}{2m}$ as a way to increase the cost contribution of outlier points, where $m$ is the number of high-dimensional datapoints. The conditional probability density $p_{j|i}$ is,

\begin{equation}p_{j|i} = \frac{\exp(-d(\mathbf{x}_i, \mathbf{x}_j)^2 / 2\sigma_i^2)}{\sum_{k\neq i}\exp(-d(\mathbf{x}_i, \mathbf{x}_k)^2 / 2\sigma_i^2)}
\label{eq3}
\end{equation}
where $d(\mathbf{x}_i, \mathbf{x}_j)$ is the distance between $\mathbf{x}_i$ and $\mathbf{x}_j$. In the low-dimensional space, Student's t-distribution is used for modeling the joint probability distribution $Q$ between embeddings, and $q_{ij}$ is defined as,

\begin{equation}
q_{ij} = \frac{(1+d(\mathbf{y}_i, \mathbf{y}_j)^2)^{-1}}{\sum_{k\neq l}(1+d(\mathbf{y}_k, \mathbf{y}_l)^2)^{-1}}
\label{eq4}
\end{equation}
where $\mathbf{y}_i$ is the corresponding low-dimensional embedding of $\mathbf{x}_i$. In t-SNE, to maintain the local similarities, the cost function to minimize is the Kullback-Leibler divergence between the probability distributions $P$ and $Q$:
\begin{equation}
\mathcal{C} = KL(P||Q) = \sum_i\sum_jp_{ij} \log{\frac{p_{ij}}{q_{ij}}}
\end{equation}

One direct extension of t-SNE is to replace Euclidean version of normal distribution and Student's t-distribution with hyperbolic normal distribution and hyperbolic Student's t-distribution. We call such a direct extension as HT-SNE. In Section \ref{sec:normal} and Section \ref{sec:student_t}, we show how to generalize the distributions to hyperbolic space.

\begin{figure*}[!t]
   \centering
\begin{tabular}{cccc}
\includegraphics[width=0.2\textwidth]{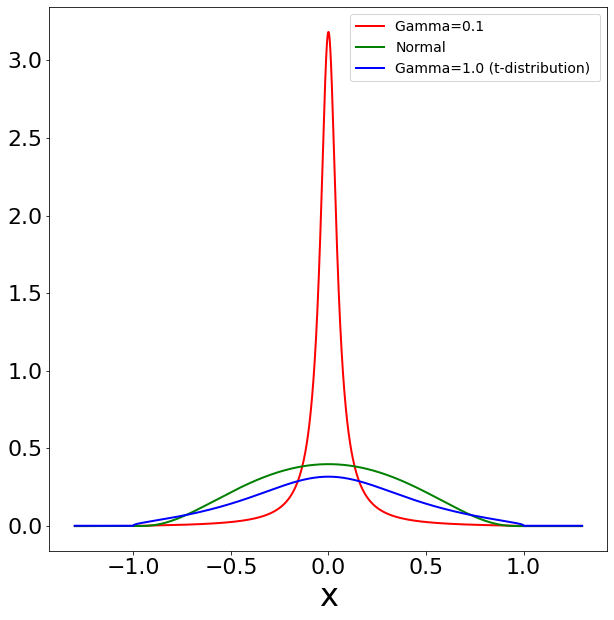} &
\includegraphics[width=0.23\textwidth]{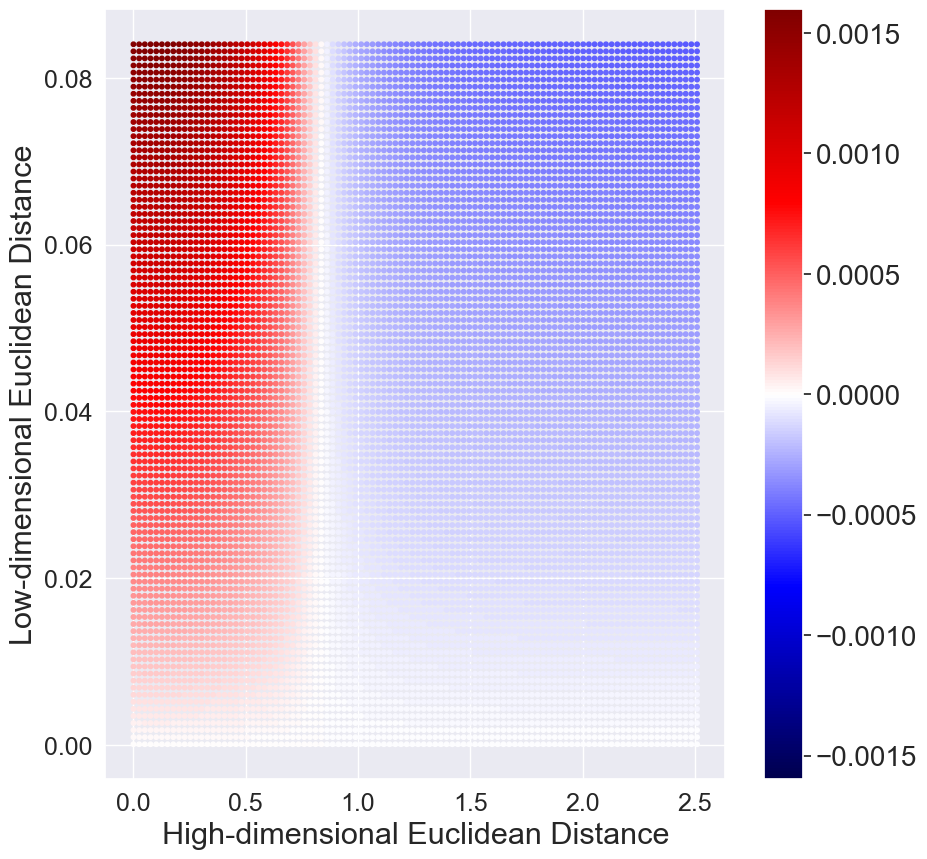} &
\includegraphics[width=0.23\textwidth]{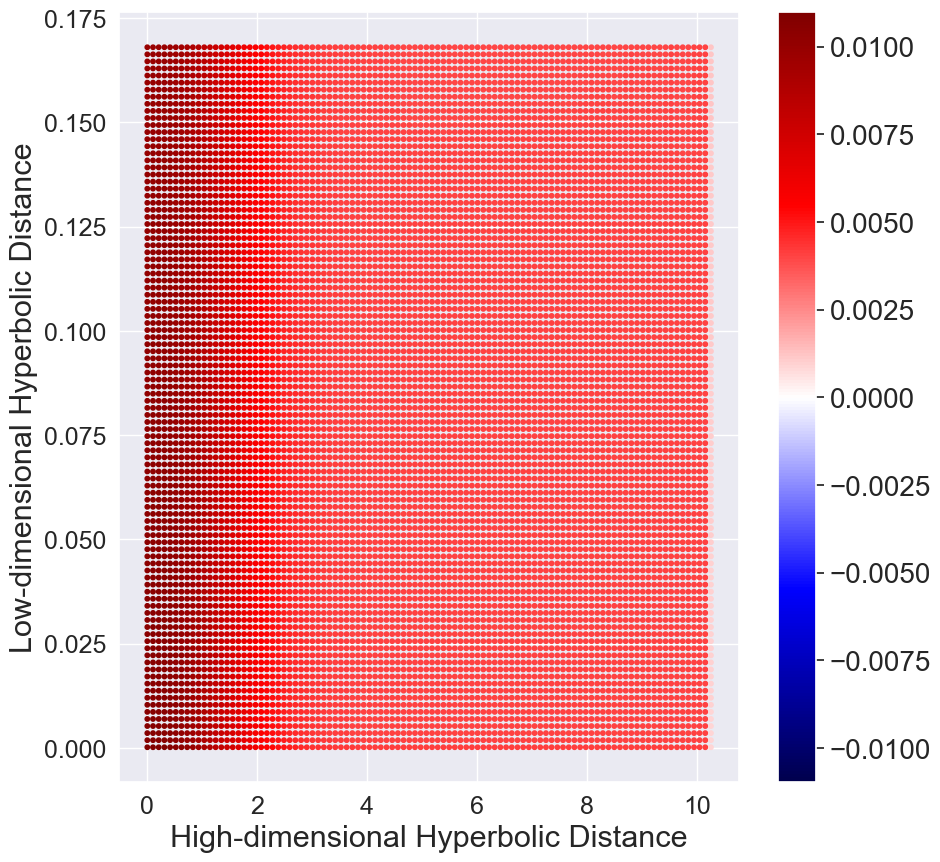} &
\includegraphics[width=0.23\textwidth]{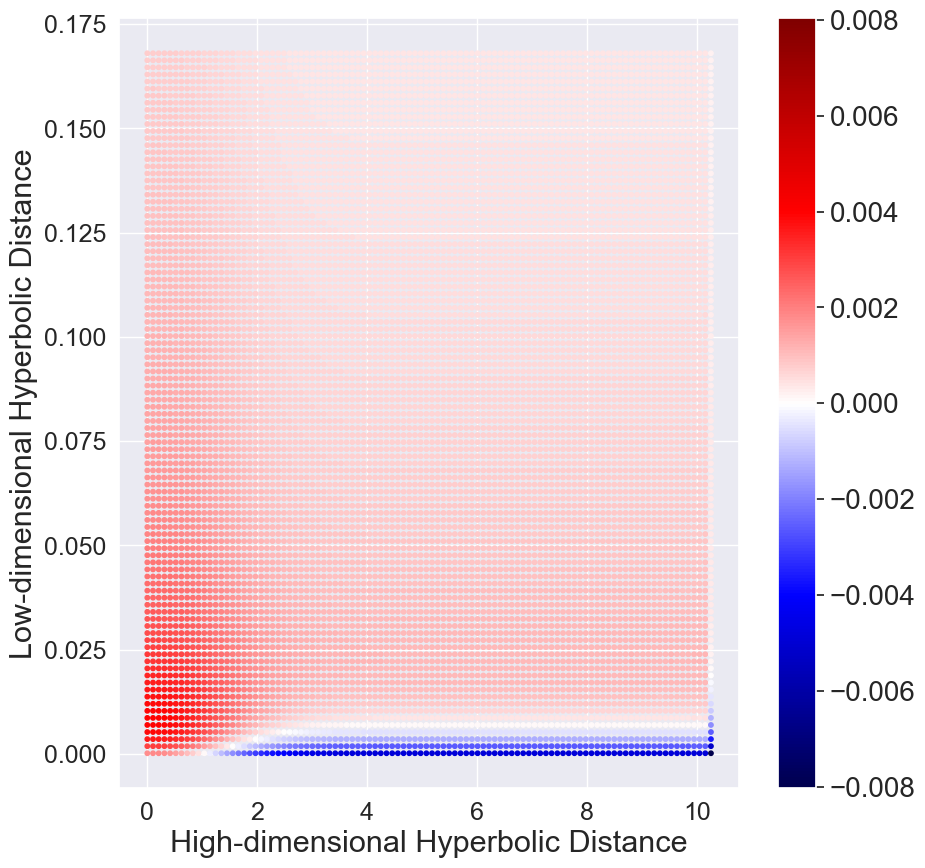} \\
    a) Normal and Cauchy Dist. & b) t-SNE & c) HT-SNE & d) CO-SNE 
\end{tabular}
    \caption{Hyperbolic Student's t-distribution is not heavy-tailed which leads to strong attraction forces in HT-SNE. a) The probability density function of the hyperbolic normal distribution, hyperbolic Cauchy distribution with $\gamma = 0.1$ and hyperbolic Cauchy distribution with $\gamma = 1.0$. Hyperbolic Student's t-distribution (hyperbolic Cauchy distribution with $\gamma = 1.0$) is not heavy-tailed. b) The gradients of the standard t-SNE as a function of low-dimensional and high-dimensional Euclidean distance. c) The gradients of the HT-SNE as a function of low-dimensional and high-dimensional hyperbolic distance. d) The gradients of CO-SNE as a function of low-dimensional and high-dimensional hyperbolic distance. There is strong repulsion when dissimilar high-dimensional datapoints are projected close.}
    \label{fig:distribution_comparison}
\end{figure*}

\subsection{Hyperbolic Normal Distribution}
\label{sec:normal}
To define the conditional probability in the high-dimensional hyperbolic space, we need to generalize the normal distribution to hyperbolic space. One natural generalization is called the Riemannian normal distribution which is the maximum entropy probability distribution given an expectation and a variance \cite{pennec2006intrinsic}. Given the Fr\'echet mean $\bm{\mu} \in \mathbb{B}^n_c$ and a dispersion parameter $\sigma > 0$, the Riemannian normal distribution is defined as,
\begin{equation}
    \mathcal{N}_{\mathbb{B}^n}(\bm{x}|\bm{\mu}, \sigma^2) = \frac{1}{\bm{Z}} \exp(-\frac{ d_{\mathbb{B}^n}(\bm{\mu}, \bm{x})^2 }{2\sigma^2})
\end{equation}
where $\bm{Z}$ is the normalization constant. There are other generalizations of the normal distribution in hyperbolic space \cite{mathieu2019continuous}. We use the Riemannian normal distribution for simplicity. Thereafter, we refer to Riemannian normal distribution as hyperbolic normal distribution.

\subsection{Hyperbolic Student's t-Distribution}
\label{sec:student_t}
 One way to define the Student's t-distribution is to express the random variable $t$ as,
\begin{equation}
    t = \frac{u}{\sqrt{v/n}}
    \label{eq: t_variable}
\end{equation}
where $u$ is a random variable sampled from a standard normal distribution and $v$ is a random variable sampled from a $\chi^2$-distribution of $n$ degrees of freedom. In particular, t-SNE adopts a Student's t-distribution with one degree of freedom and the probability density function is defined as, 
\begin{equation}
        f(t; t_0) = \frac{1}{ \pi (1+(t-t_0)^2)}
\end{equation}
To extend the Student's t-distribution to hyperbolic space, we derive the probability density function as,
\begin{equation}
        f_{\mathbb{B}^n}(t; t_0) = \frac{1}{ \pi (1+d_{\mathbb{B}^n}(t, t_0)^2)}
\end{equation}
The details can be found in the Supplementary.

\subsection{Hyperbolic t-Distribution is Not Enough}
\label{sec:not_heavy}
The motivation to use Student's t-distribution in the standard t-SNE is that Student's t-distribution has heavier tails than the normal distribution. This causes a repulsion force between dissimilar high-dimensional datapoints which helps mitigate the ``Crowding Problem" \cite{van2008visualizing}. However, hyperbolic Student's t-distribution is not heavy-tailed since the hyperbolic distance grows exponentially fast. 

Figure \ref{fig:distribution_comparison} plots the hyperbolic normal distribution and hyperbolic Student's t-distribution. It can be noted that the tails of hyperbolic Student's t-distribution diminish as fast as hyperbolic normal distribution. In contrast, the standard Euclidean Student's t-distribution has heavier tails than the normal distribution. The consequence is that the repulsion force and the attraction force with hyperbolic Student's t-distribution behave drastically different from that of using Euclidean Student's t-distribution.

Similar to \cite{van2008visualizing}, we plot the gradients between two low-dimensional embeddings $\mathbf{y}_i$ and $\mathbf{y}_j$ as a function of their pairwise Euclidean (hyperbolic) distance in the low-dimensional Euclidean (hyperbolic) and the pairwise Euclidean (hyperbolic) distance of the corresponding datapoints in the high-dimensional Euclidean (hyperbolic) space. Positive values indicate attraction and negative values indicate repulsion between the embeddings $\mathbf{y}_i$ and $\mathbf{y}_j$ respectively. Here we only consider the initial stage when all the low-dimensional embeddings are close. The results are shown in Figure \ref{fig:distribution_comparison}. We have two observations. First, for the standard t-SNE, there is strong repulsion when dissimilar high-dimensional datapoints are projected closely. Second, with the hyperbolic Student's t-distribution in HT-SNE, there is little or no repulsion force which causes the low-dimensional embeddings tend to cluster together. 

To remedy this issue, we seek to replace the hyperbolic Student's t-distribution to create more repulsion forces between dissimilar high-dimensional datapoints.  Consider the hyperbolic Cauchy distribution which has the probability density function,
\begin{equation}
    f(t; t_0, \gamma) = \frac{1}{\pi \gamma} [ \frac{\gamma^2}{ d_{\mathbb{B}^n}(t, t_0)^2 + \gamma^2} ]
\end{equation}
where $\gamma$ is the scale parameter. Notice that the Student's t-distribution is a special case of Cauchy distribution with $\gamma = 1.0$. With a small $\gamma$, Cauchy distribution has a higher peak (Figure \ref{fig:distribution_comparison} (a)). Thus, if the high-dimensional datapoints are modeled with close low-dimensional embeddings, the density value is much larger than the corresponding density in the high-dimensional space. This produces a strong repulsion when dissimilar high-dimensional datapoints are modeled with close low-dimensional embeddings. More details can be found in the Supplementary. We show the corresponding gradients in Figure \ref{fig:distribution_comparison} (d). We can observe that there is a strong repulsion force when high-dimensional dissimilar datapoints are projected closely in CO-SNE. Such repulsion is particularly important in the initial stage of training since all the low-dimensional embeddings are close.

\subsection{Distance Loss}
\label{sec:rcl}
Hyperbolic space can be naturally used for embedding tree structured data. The datapoint which is close to the center of the Poincar\'e ball can be viewed as the root node and datapoints which are close to the boundary of the Poincar\'e ball can be viewed as leaf nodes. The criterion of t-SNE cannot preserve the tree-structure of the high-dimensional datapoints in the low-dimensional space. One natural way to characterize the level of the datapoint in the underlying tree is the norm of the datapoint. Thus, we attempt to keep the norm invariant after the high-dimensional datapoint is projected. This leads to the following loss function which minimizes the difference between the norms of the high-dimensional datapoint $\mathbf{x}_i$ and the corresponding low-dimensional embedding $\mathbf{y}_i$,
\begin{equation}
    \mathcal{H} = \frac{1}{m} \sum_{i=1}^m ( \lVert \mathbf{x}_i\rVert ^2 - \lVert \mathbf{y}_i \rVert ^ 2)^2
\end{equation}
With the distance loss, we can preserve the global hierarchy of the high-dimensional hyperbolic embeddings. 

\begin{table}[t]
\scriptsize
    \centering
    \begin{tabular}{l|c| c |c}
 \toprule
         & Metric & High/low-dimensional Dist.  &  Losses   \\
    \midrule
              \rowcolor{Gray}
       t-SNE  & Euclidean  & Normal/t-distribution & KL-div  \\ 
    \midrule
       HT-SNE  & Hyperbolic  & Normal/t-distribution & KL-div  \\
    \midrule
       CO-SNE & Hyperbolic & Normal/Cauchy & KL-div + Distance \\
    \bottomrule
    \end{tabular}
    \caption{Our CO-SNE extends t-SNE by adopting hyperbolic normal distribution and hyperbolic Cauchy distribution. Compared with t-SNE, CO-SNE assumes the high-dimensional and low-dimensional space are hyperbolic. t-SNE cannot maintain the global hierarchy of the hyperbolic embeddings. Compared with HT-SNE, CO-SNE adopts hyperbolic Cauchy distribution and an additional distance loss. HT-SNE cannot push dissimilar high-dimensional points away in the low-dimensional space. }
    \label{tab:sne_comparison}
\end{table}

\subsection{Optimization of CO-SNE}
\label{sec:criterion}

\noindent \textbf{Criterion.} The criterion of CO-SNE is composed of the KL-divergence for maintaining local similarity and the distance loss for maintaining the global hierarchy,
\begin{equation}
    \mathcal{L} = \lambda_1 \mathcal{C} + \lambda_2 \mathcal{H}
\end{equation}
where $\lambda_1$ and $\lambda_2$ are hyperparameters. Table \ref{tab:sne_comparison} shows a comparison of t-SNE, HT-SNE and CO-SNE.

\noindent \textbf{Gradients.} The cost function is optimized by gradient descent with respect to the low-dimensional embeddings. The gradient of the KL-divergence $\mathcal{C}$ with respect to $\mathbf{y}_i$ is given by,
\begin{equation}
\begin{split}
\frac{\delta \mathcal{C}}{\delta \mathbf{y}_i} & =\sum_j \frac{\delta \mathcal{C}}{\delta d_{\mathbb{B}^n}(\mathbf{y}_i, \mathbf{y}_j)}\frac{\delta d_{\mathbb{B}^n}(\mathbf{y}_i, \mathbf{y}_j)}{\delta \mathbf{y}_i} \\
&= 2\sum_j(p_{ij} - q_{ij})(1+d_{\mathbb{B}^n}(\mathbf{y}_i, \mathbf{y}_j)^2)^{-1}\frac{\delta d_{\mathbb{B}^n}(\mathbf{y}_i, \mathbf{y}_j)}{\delta \mathbf{y}_i}     
\end{split}
\label{eq:local_gradient}
\end{equation}
The partial gradient of the distance with respect to the low dimensional embedding $\mathbf{y}_i$ is given by 
\begin{equation}
\small
\frac{\delta d_{\mathbb{B}^n}(\mathbf{y}_i, \mathbf{y}_j)}{\delta \mathbf{y}_i} = \frac{4}{\beta\sqrt{\gamma^2 -1}}\left( \frac{||\mathbf{y}_j||^2 - 2\langle \mathbf{y}_i, \mathbf{y}_j\rangle + 1}{\alpha^2}\mathbf{y}_i - \frac{\mathbf{y}_j}{\alpha}\right)
\end{equation}
where $\alpha=1-||\mathbf{y}_i||^2$, $\beta=1-||\mathbf{y}_j||^2$, $\gamma = 1 + \frac{2}{\alpha\beta}||\mathbf{y}_i-\mathbf{y}_j||^2$.
We use the Riemannian stochastic gradient descent \cite{bonnabel2013stochastic} for the KL-divergence $\mathcal{C}$. The gradient of the distance loss $\mathcal{H}$ with respect to $\mathbf{y}_i$ is computed as,
\begin{equation}
    \frac{\delta \mathcal{H}}{\delta \mathbf{y}_i} = -4( \lVert \mathbf{x}_i\rVert ^2 - \lVert \mathbf{y}_i \rVert ^ 2)\mathbf{y}_i
\end{equation}
We constrain the embeddings to the Poincar\'e ball after each update as in \cite{nickel2017poincar}.

\noindent \textbf{Stagewise Training.} We found the following training strategy is useful for producing better visualization in practice. We train the low-dimensional embeddings only with the local similarity loss for the first 500 iterations. Then we add the distance loss. The stopping criterion is the same as in t-SNE \cite{van2008visualizing}. The reason of the strategy is that it is hard to move the points when they are approaching the boundary of the Poincar\'e ball.

\section{Experiments and Results}

\noindent \textbf{Baselines.} We compare CO-SNE with 4 baselines,
\begin{itemize}
    \item The standard t-SNE \cite{van2008visualizing}: this is the standard t-SNE which adopts Euclidean distance for computing the similarities in the high-dimensional space and the low-dimensional space. 
    
    \item  Principal Component Analysis (PCA) \cite{jolliffe2016principal}: PCA is a commonly used dimensionality reduction method which attempts to maintain the maximal variation of the data. However, as a linear dimensionality method, PCA cannot reduce high-dimensional data to two dimensions in a meaningful way  \cite{arora2018analysis}.

    \item HoroPCA \cite{chami2021horopca}: HoroPCA is a recently proposed extension of PCA on hyperbolic space. HoroPCA proposed to parameterize geodesic subspaces by ideal points in the Poincar\'e ball. HoroPCA also generalized the notion of projection and the objective function of PCA to hyperbolic space. HoroPCA can be used as a data whitening method for hyperbolic data and as a visualization method as well.
    
    \item UMAP \cite{mcinnes2018umap}: UMAP is a recently proposed dimensionality reduction and visualization method based on the ideas from Riemannian geometry and topology. UMAP is competitive with t-SNE and can better preserve the global structure of the data. 
\end{itemize}

        \noindent \textbf{Tasks.} We consider 4 datasets to validate the effectiveness of CO-SNE: 1) synthetic datasets sampled from a mixture of hyperbolic normal distributions, 2) cellular differentiation data, 3) supervised hyperbolic embeddings and 4) unsupervised hyperbolic embeddings.

\noindent \textbf{Implementation Details.} For PCA and the standard t-SNE, we use the implementation from \cite{scikit-learn}. For UMAP, we use the implementation from \cite{mcinnes2018umap-software}. For HoroPCA, we use the implementation from \url{https://github.com/HazyResearch/HoroPCA} which is provided by the original authors. We implement CO-SNE based on t-SNE by modifying the way of computing similarities and the optimization procedure.

\noindent \textbf{Hyperparameters.} For PCA, UMAP and HoroPCA, we use the default hyperparameters. For SNE-based methods, we initialize the low-dimensional embedding using a normal distribution with mean 0.01 and unit variance. The training follows the standard setups as in \cite{scikit-learn}. For CO-SNE, the scaling parameter $\gamma$ is set to 0.1. The hyperparameter $\lambda_1$ is usually set to 10.0 and the hyperparameter $\lambda_2$ is usually set to 0.01.

\begin{figure}
    \centering
    \setlength{\tabcolsep}{0pt}
\begin{tabular}{ccc}
    \includegraphics[width=0.3\linewidth]{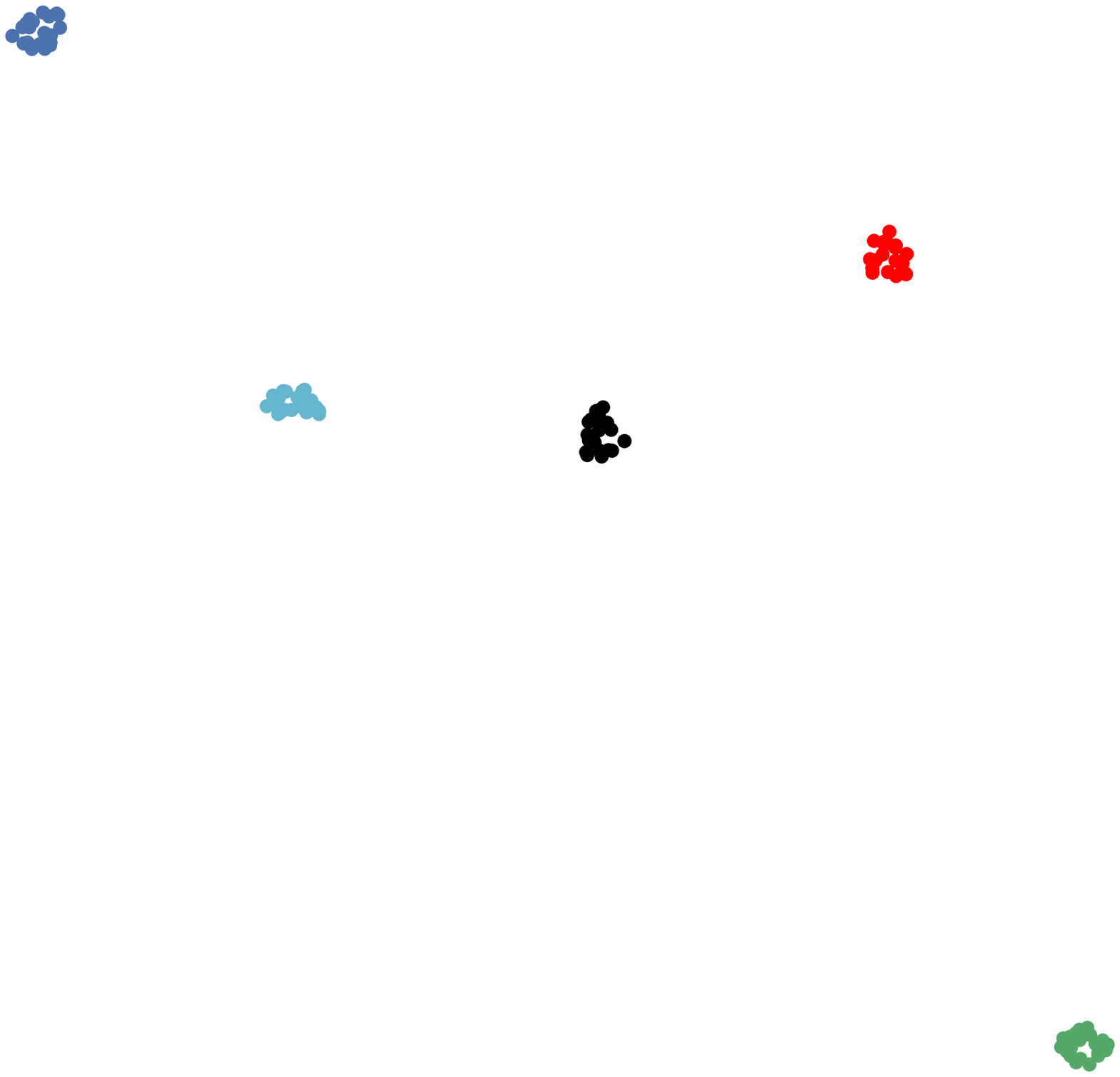} & \includegraphics[width=0.3\linewidth]{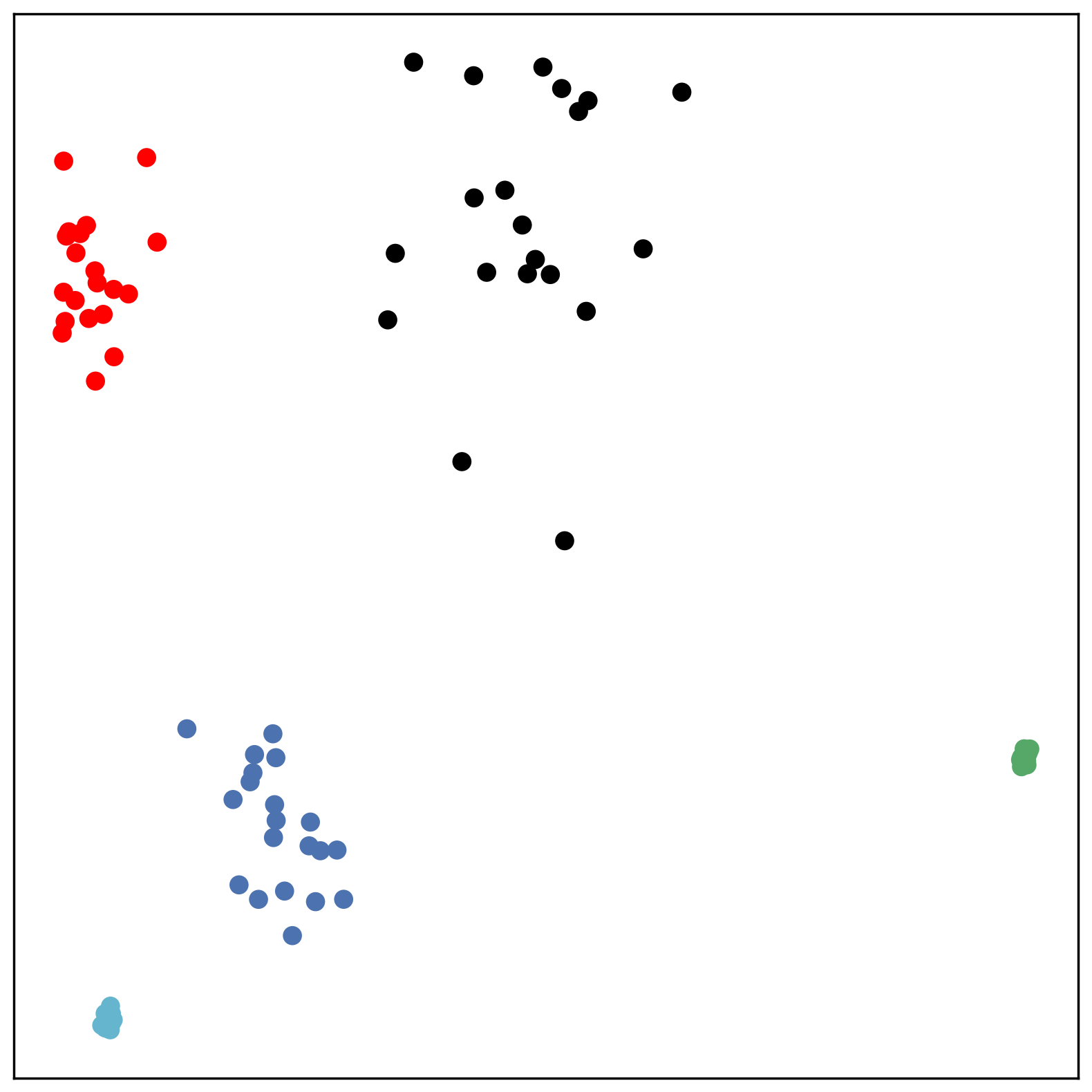} & \includegraphics[width=0.3\linewidth]{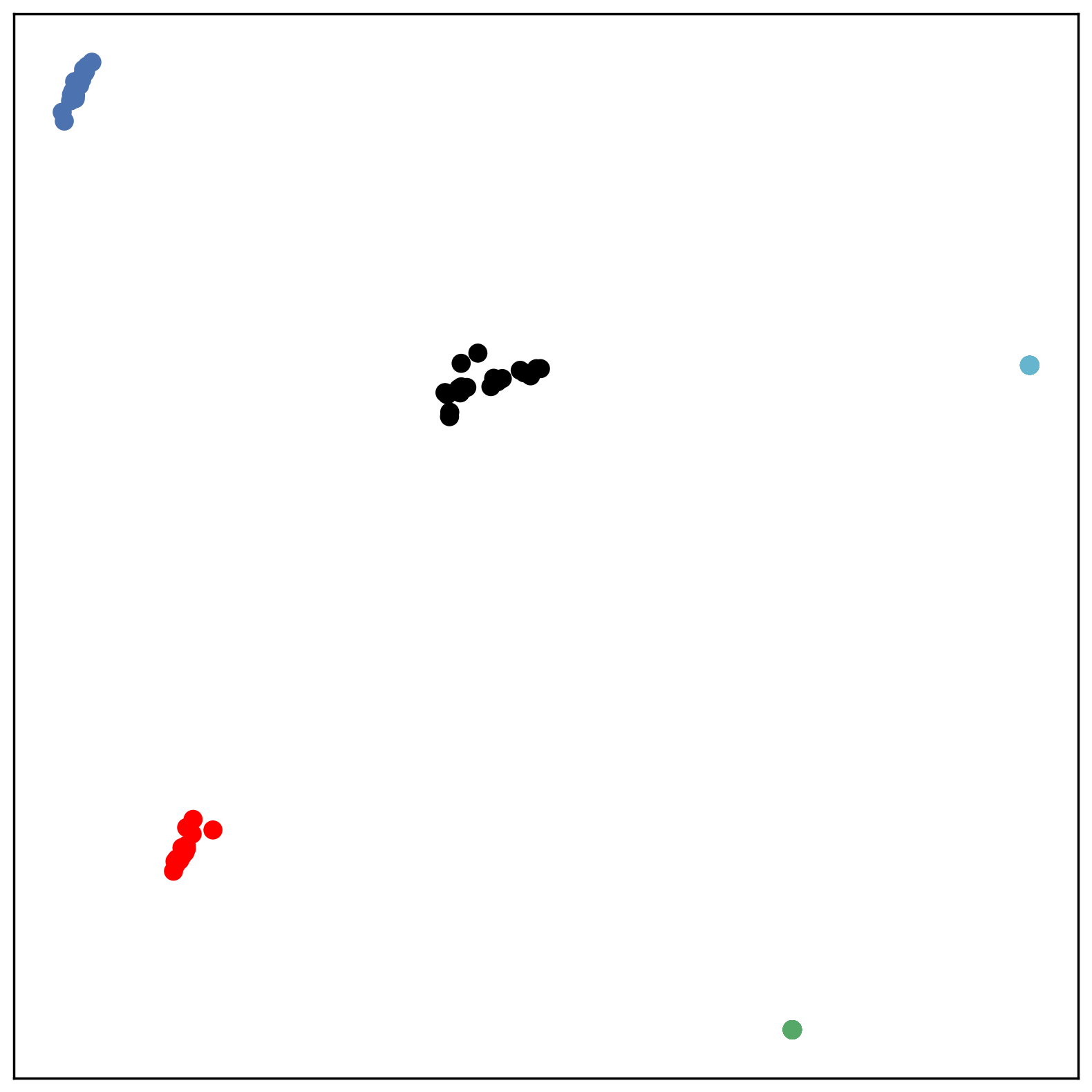} \\
   UMAP & PCA & t-SNE \\
    \includegraphics[width=0.3\linewidth]{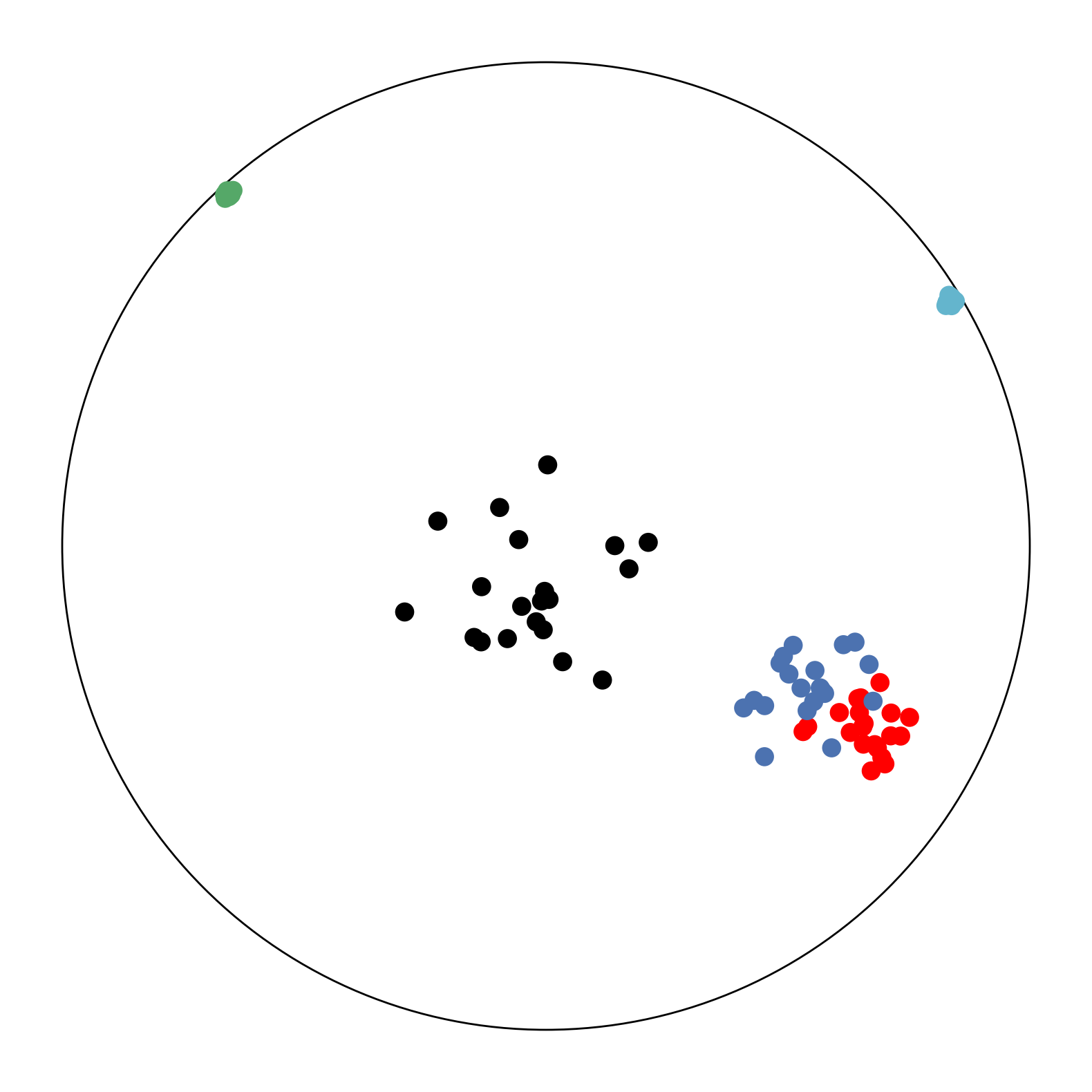} & \includegraphics[width=0.3\linewidth]{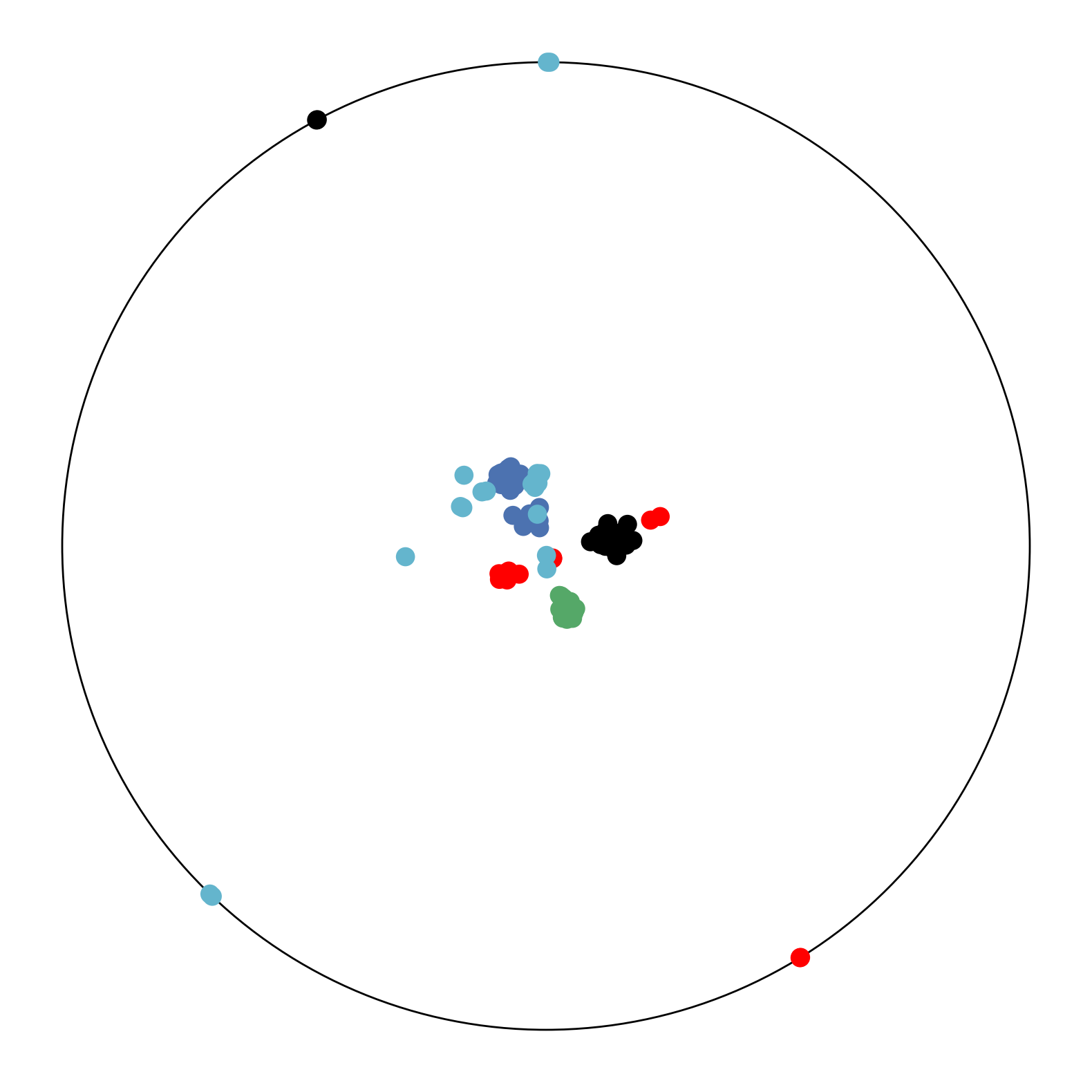} & \includegraphics[width=0.3\linewidth]{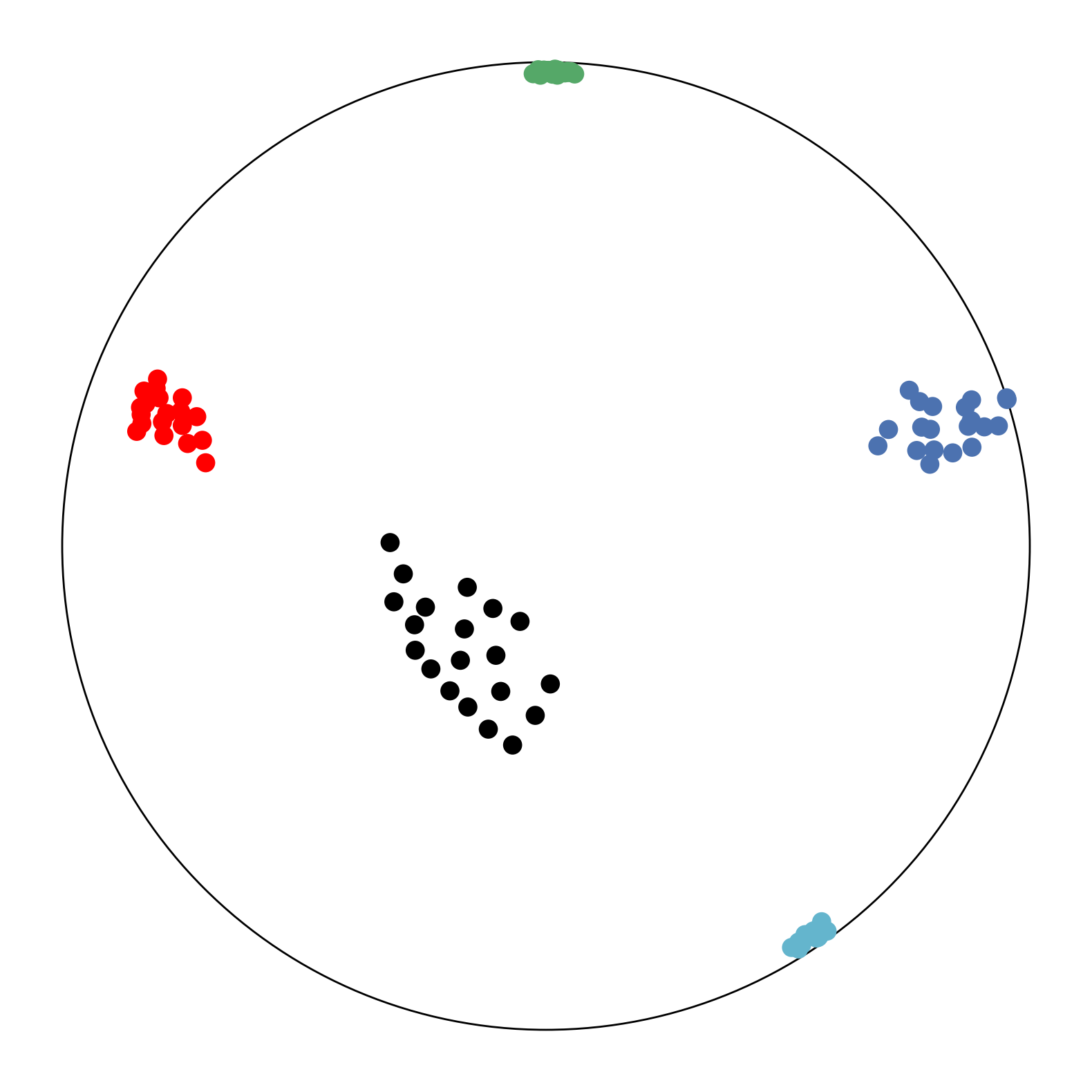} \\
      HoroPCA & HT-SNE & CO-SNE \\
\end{tabular}
    \caption{The projection of high-dimensional hyperbolic datapoints sampled from a mixture of hyperbolic normal distributions in a two-dimensional space with different methods. CO-SNE produces two-dimensional hyperbolic embeddings which preserve the hierarchical and similarity structure of the high-dimensional hyperbolic datapoints.}
    \label{fig:all_projections}
\end{figure}

\subsection{Synthetic Point Clusters}
\label{sec:synthetic}
We first use a synthetic dataset to validate the efficacy of CO-SNE.  We randomly generate five point clusters of 20 points each in the 5D hyperbolic space.  Each follows a hyperbolic normal distribution with the unit variance and the mean located on a different axis.  The first and  second means are close to the origin, at \textcolor{red}{[0.1, 0, 0, 0, 0]} and \textcolor{blue}{[0, -0.2, 0, 0, 0]} respectively, whereas the third and fourth means are far from and equi-distance to the origin, at 
\textcolor{green}{[0, 0, 0.9, 0, 0]} and \textcolor{cyan}{[0, 0, 0, -0.9, 0]} respectively. The last mean is right at the origin \textcolor{black}{[0, 0, 0, 0, 0]}. Figure \ref{fig:all_projections} shows the 2D  visualization results of these 5D points by different methods.

CO-SNE produces a much better visualization compared with the baselines. With CO-SNE, the projected two-dimensional hyperbolic embeddings can preserve the local similarity structure and the global hierarchical structure of high-dimensional datapoints well. Also, CO-SNE can prevent the high-dimensional datapoints from being projected too close which usually happens with  Euclidean distance based  methods. Note that HT-SNE does not have not enough repulsion to push close low-dimensional embeddings apart. We provide a detailed analysis on the drawbacks of each baseline for visualizing high-dimensional hyperbolic datapoints.
\begin{itemize}
    \item Standard t-SNE:  Euclidean distances are used for computing similarities in the high-dimensional space.  Hyperbolic distances grow much faster than Euclidean distances. For high-dimensional hyperbolic datapoints which are close to boundary of the Poincar\'e ball, the standard t-SNE wrongly underestimates the distance between them. As a consequence, the standard t-SNE would take dissimilar high-dimensional data points as neighbors. The resulting low-dimensional embeddings would collapse into one point which leads to poor visualization. In summary, t-SNE cannot preserve the global hierarchy of the hyperbolic data.
    
    \item PCA and HoroPCA: as mentioned above, PCA and HoroPCA are linear dimensionality reduction methods which are not generally suitable for visualization in a two-dimensional space. Both PCA and HoroPCA cannot preserve local similarity of the hyperbolic data. 
    \item UMAP: UMAP suffers from the same issue as t-SNE since Euclidean distance is used for computing high-dimensional similarities. 
\end{itemize}

We mainly compare CO-SNE with HoroPCA since HoroPCA is specifically designed for hyperbolic data.

\begin{figure}[!t]
   \centering
\setlength{\tabcolsep}{0pt}
\begin{tabular}{c}
\includegraphics[width=0.38\textwidth]{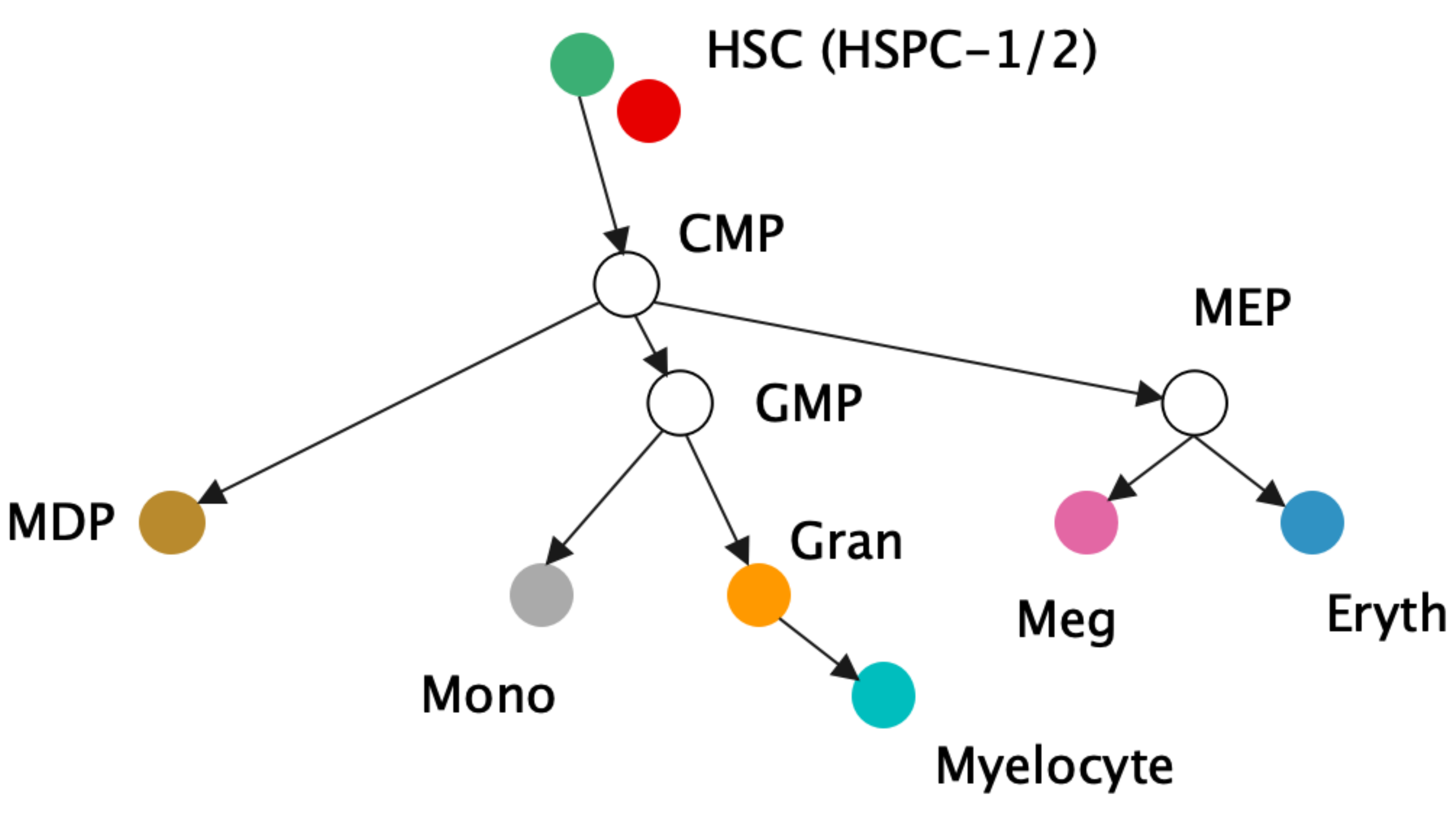} \\
a) Canonical hematopoetic cell lineage tree \cite{klimovskaia2020poincare} \\
\setlength{\tabcolsep}{0pt}
\begin{tabular}{@{\hspace{-6pt}}cc@{}}
\includegraphics[height=0.18\textheight]{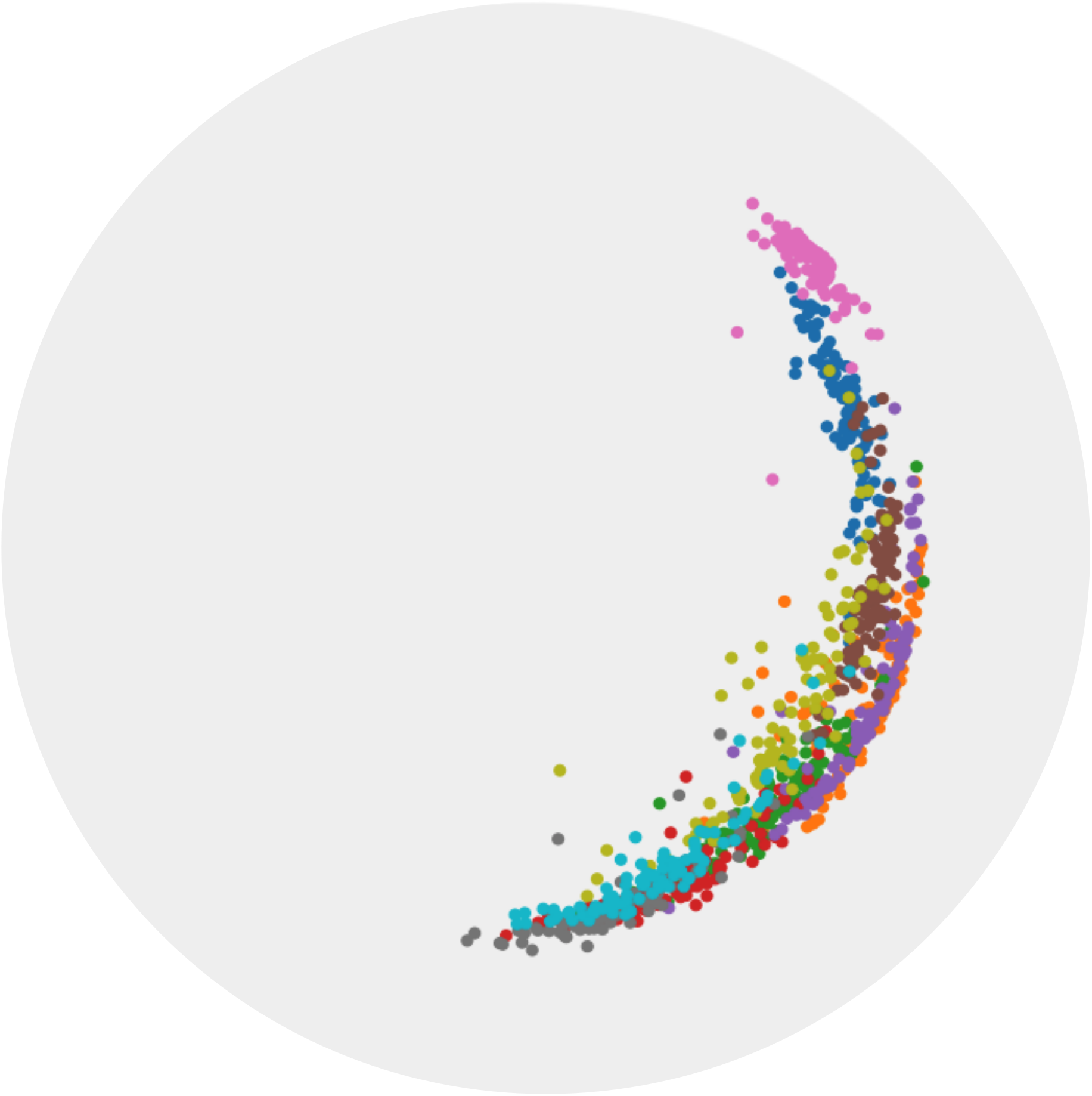} &
\includegraphics[height=0.18\textheight]{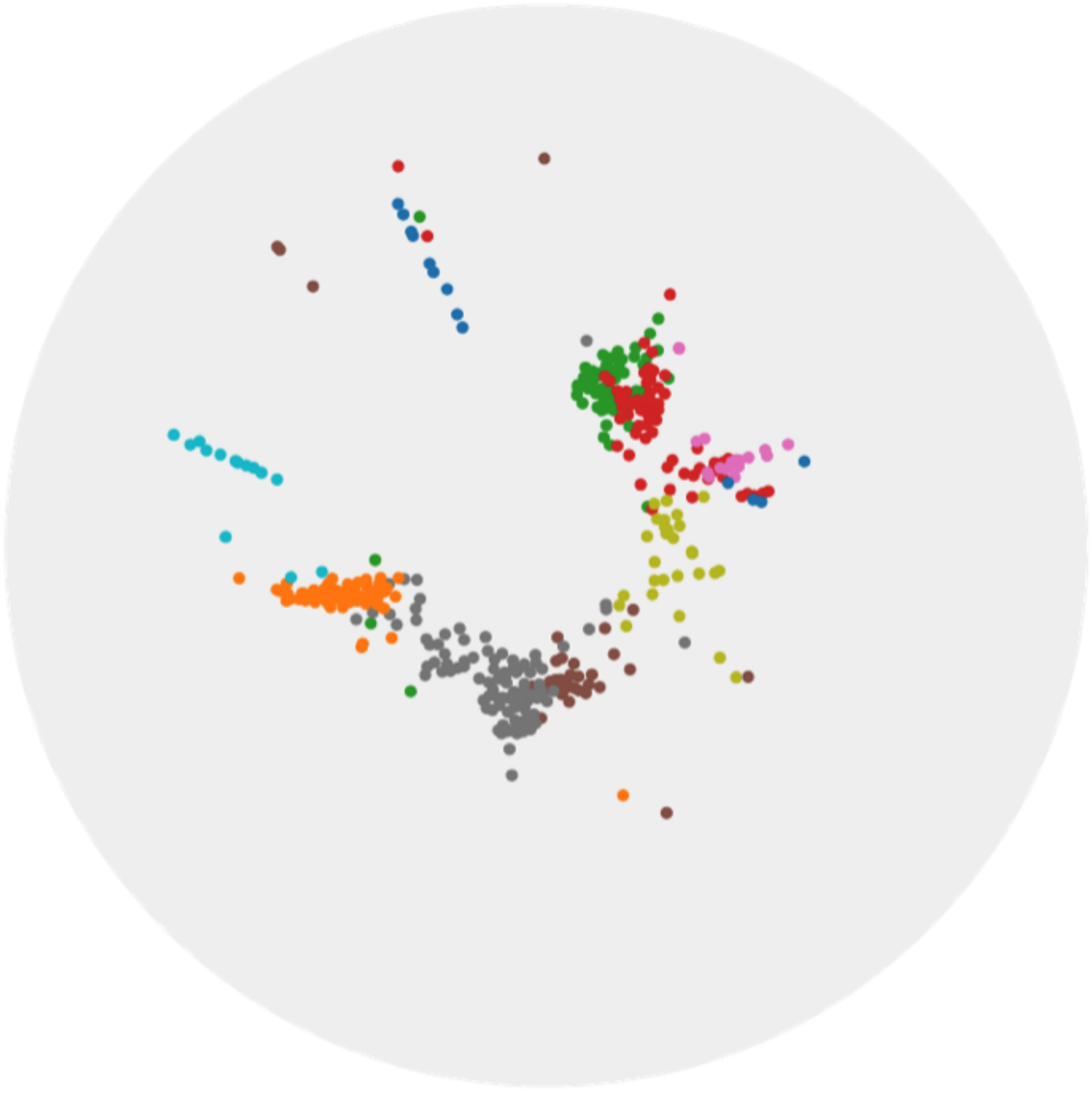}  \\
b) HoroPCA &
c) CO-SNE\\
\end{tabular}\\
\end{tabular}
    \caption{CO-SNE produces better visualization of the high-dimensional biological datapoints than HoroPCA which captures the hierarchical and similarity structure in the data. a) The hierarchy of the original data from \cite{klimovskaia2020poincare}. b) The two-dimensional embeddings generated by HoroPCA. 
    c) The two-dimensional embeddings generated by CO-SNE.}
    \label{fig:bio_embeddings}
\end{figure}

\subsection{Biological Dataset}


Biological data can reveal naturally occurring hierarchies, such as in single cell RNA sequencing data \cite{klimovskaia2020poincare}. In \cite{klimovskaia2020poincare}, they analyze cellular differentiation data, the transition of immature cells to specialized types. The immature cells can be viewed as the root of the tree and can branch off into several different types of cells, creating hierarchical data of cells in different states of progress in the transition process. One dataset we adapt from \cite{klimovskaia2020poincare} is the mouse myelopoiesis dataset presented by \cite{olsson2016singlecell}, where there are 532 cells of 9 types. Two of the types, HSPC-1 and HSPC-2, form the root of the hierarchy, while megakaryocytic (Meg), erythrocytic (Eryth), monocyte-dendritic cell precursor (MDP), monocytic (Mono), and myelocyte (myelocytes and metamyelocytes) type cells are states father from the root. Granulocytic (Gran) cells are a precursor to myelocytes and multi-lineage primed (Multi-Lin) cells are in an intermediate state. 

The data has originally 382 dimensions (noisy) and like \cite{klimovskaia2020poincare} we first reduce it to 20 dimensions via PCA to reduce the noises. We then scale the data to fit within the Poincar\'e ball and run CO-SNE and HoroPCA to produce two-dimensional hyperbolic embeddings. Noted that this dataset is not centered. The results are shown in Figure \ref{fig:bio_embeddings}. The proposed low-embeddings by CO-SNE capture the hierarchical structure in the original data.

\begin{figure}[!t]
   \centering
    \setlength{\tabcolsep}{0pt}
\begin{tabular}{cc}
   \includegraphics[height=0.18\textheight]{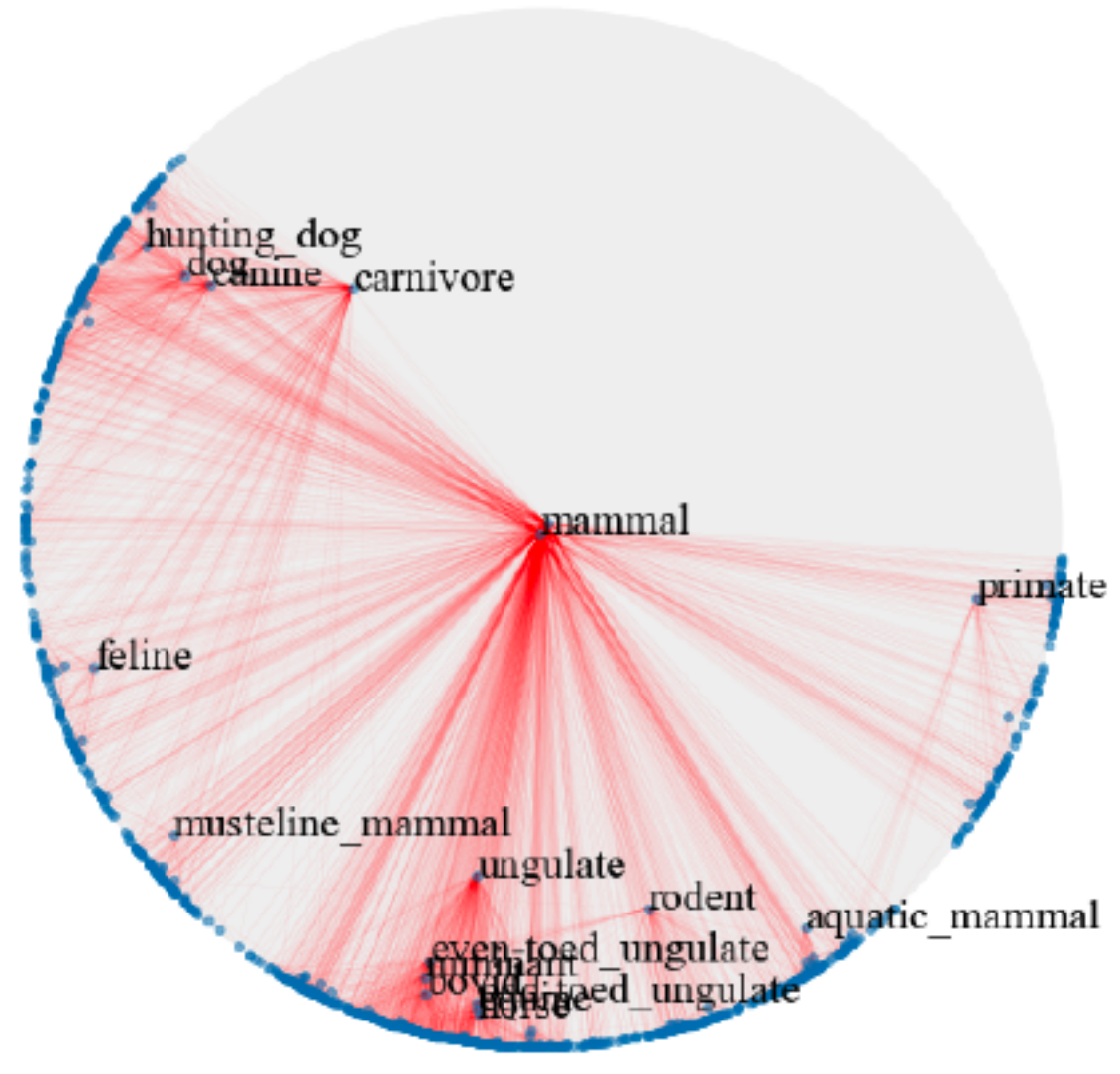} &
   \includegraphics[height=0.18\textheight]{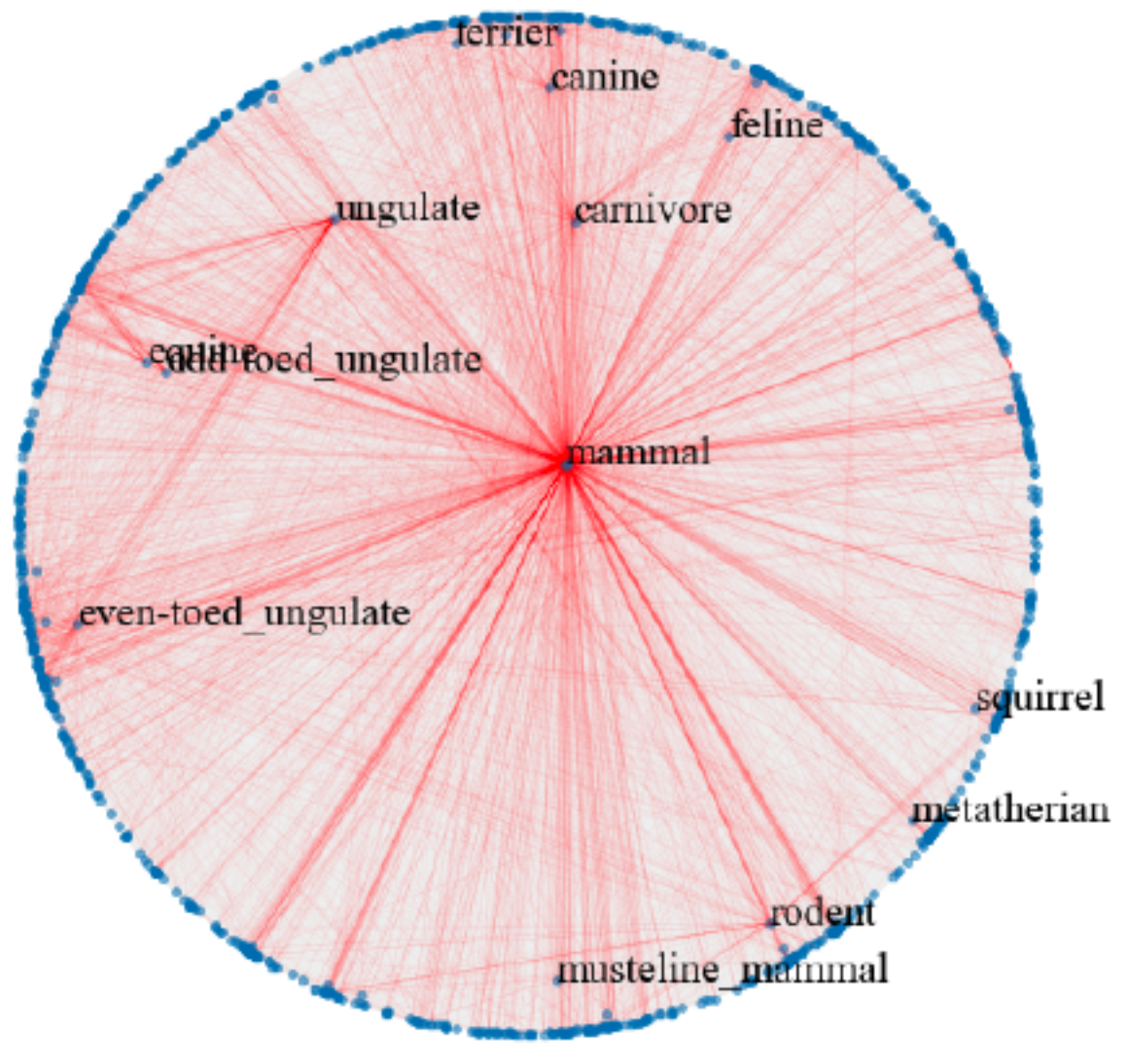} \\
   a) HoroPCA & b) CO-SNE
\end{tabular}
    \caption{CO-SNE produces better visualization of high-dimensional Poincar\'e word embeddings than HoroPCA. a) The two-dimensional Poincar\'e word embeddings generated from HoroPCA. b) The two-dimensional Poincar\'e word embeddings generated from CO-SNE. In the embeddings generated by CO-SNE, the word $feline$ and $canine$ are close to $canivore$, which is not the case in HoroPCA.}
    \label{fig:word_embeddings}
\end{figure}

\vspace{-0.15cm}
\subsection{Hierarchical Word Embeddings}
Hyperbolic space has been used to embed hierarchical representations of symbolic data. In \cite{nickel2017poincar}, the authors adopt hyperbolic space for embedding taxonomies, in particular, the transitive closure of WordNet noun hierarchy \cite{miller1995wordnet}. As shown in \cite{nickel2017poincar}, higher-dimensional hyperbolic embeddings often lead to better representations, but they are harder to visualize. Following \cite{nickel2017poincar}, we embed the hypernymy relations of the mammals subtree of WordNet in hyperbolic space. We use the open source implementation provided by \cite{nickel2017poincar} to train the ten-dimensional embeddings. We use HoroPCA and CO-SNE to visualize the learned embeddings in a two-dimensional hyperbolic space.

Figure \ref{fig:word_embeddings} shows that compared with HoroPCA, CO-SNE can better preserve the hierarchical and similarity structure of the high-dimensional datapoints. For example, the word $feline$ and $canine$ are close to $canivore$ in the CO-SNE embeddings, which is not the case in HoroPCA. The embeddings produced by CO-SNE also more resemble the two-dimensional embeddings as shown in Figure 2b of \cite{nickel2017poincar}.
 \begin{figure}[!t]
    \centering
    \setlength{\tabcolsep}{0pt}
\begin{tabular}{cc}
   \includegraphics[height=0.18\textheight]{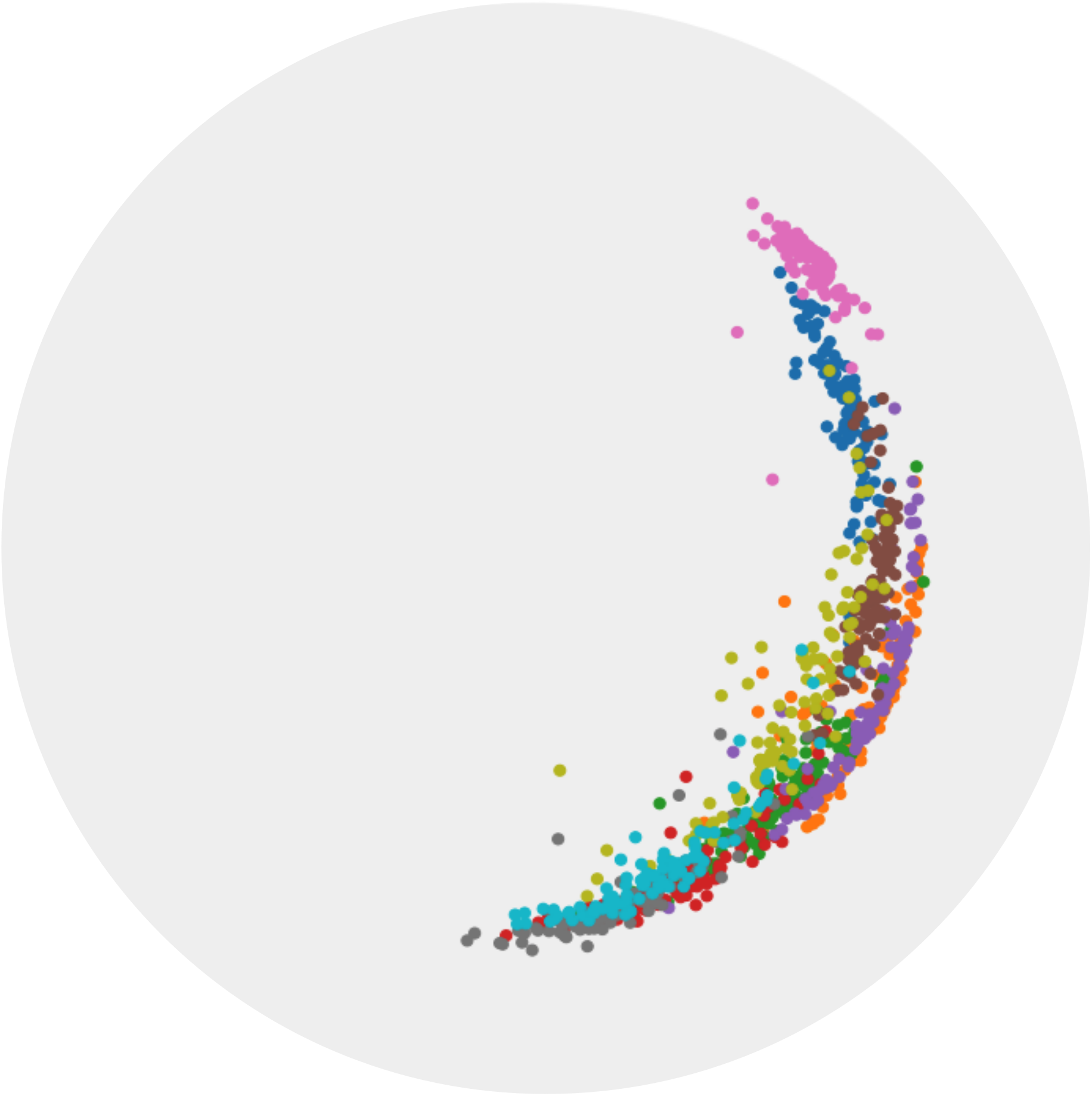} &
   \includegraphics[height=0.18\textheight]{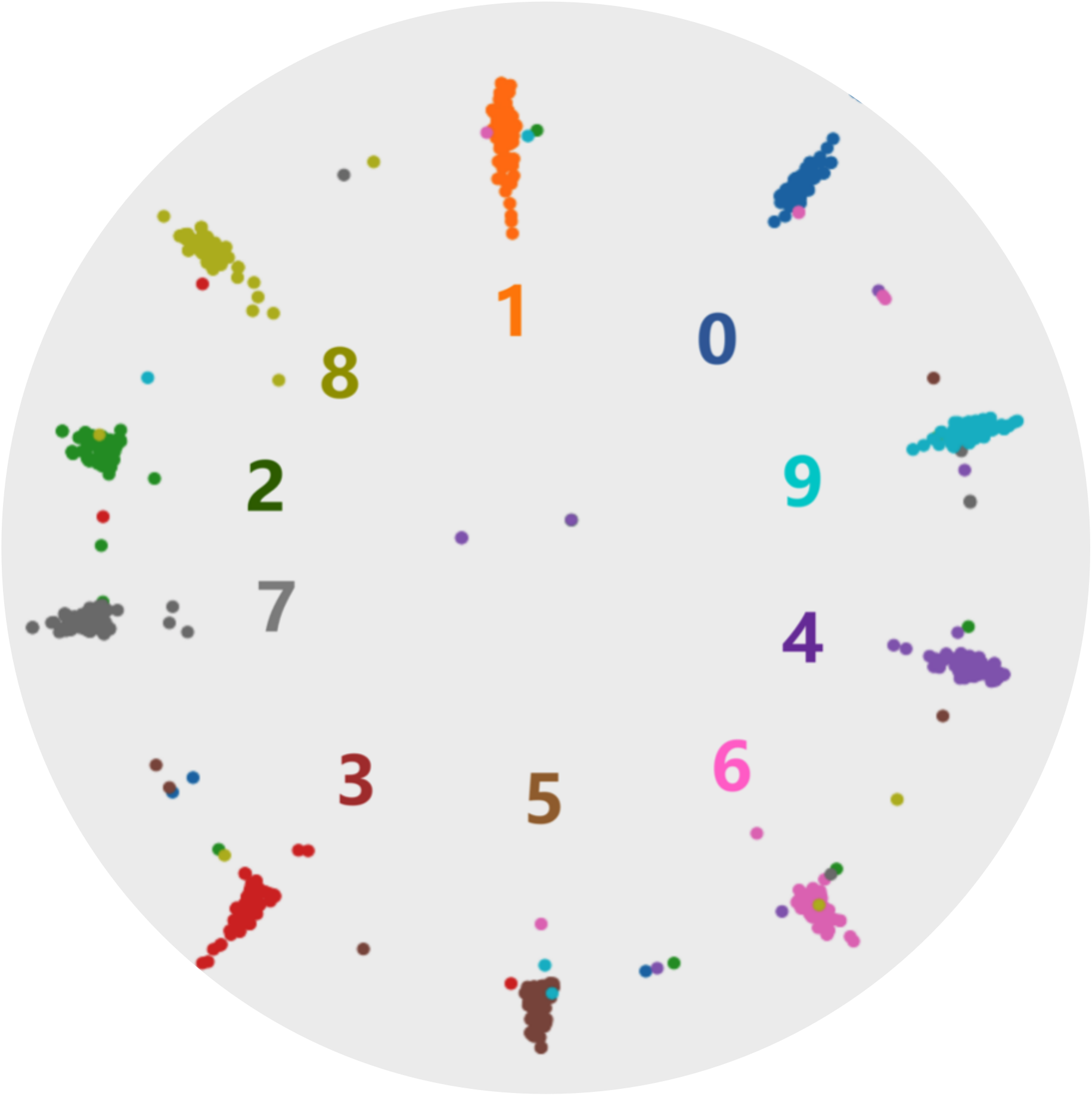} \\
   a) HoroPCA & b) CO-SNE
\end{tabular}
    \caption{CO-SNE produces better visualization of hyperbolic neural networks' (HNNs) features than HoroPCA. a) The visualization produced by HoroPCA. b) The visualization produced by CO-SNE. In CO-SNE, the classes are well separated and have a clear hierarchical structure. }
    \label{fig:hnn_result}
\end{figure}

\subsection{Features of Hyperbolic Neural Networks}
 \label{sec:d_reduction}
 
 We apply CO-SNE to visualize the embeddings produced by hyperbolic neural networks for supervised image classification. We train a hyperbolic neural network (HNN) \cite{ganea2018hyperbolic} with feature clipping \cite{guo2021free} on MNIST. The clipping value is 1.0 and the feature dimension is 64. We use HoroPCA and CO-SNE to reduce the dimensionality of the features to two. The full test set of MNIST cannot be used due to the out-of-memory issue of HoroPCA. So for each class, we randomly sample 100 images.
 
Figure \ref{fig:hnn_result} shows the two-dimensional embeddings generated by HoroPCA and CO-SNE. The visualization produced by CO-SNE is significantly better the visualization produced by HoroPCA. In CO-SNE, the classes are well separated and have a clear hierarchical structure. We further train hyperbolic classifiers \cite{ganea2018hyperbolic} on the frozen two-dimensional features. We use a learning rate of 0.001 and the number of epoch is 10. The accuracy of the features generated by HoroPCA is 30.2\% while the accuracy of the features generated by CO-SNE is 61.2\%. This implies that low-dimensional features generated by CO-SNE are more separated and respect the structure of original high-dimensional embeddings.

\begin{figure}[!t]
   \centering
       \setlength{\tabcolsep}{0pt}
\begin{tabular}{cc}
   \includegraphics[height=0.18\textheight]{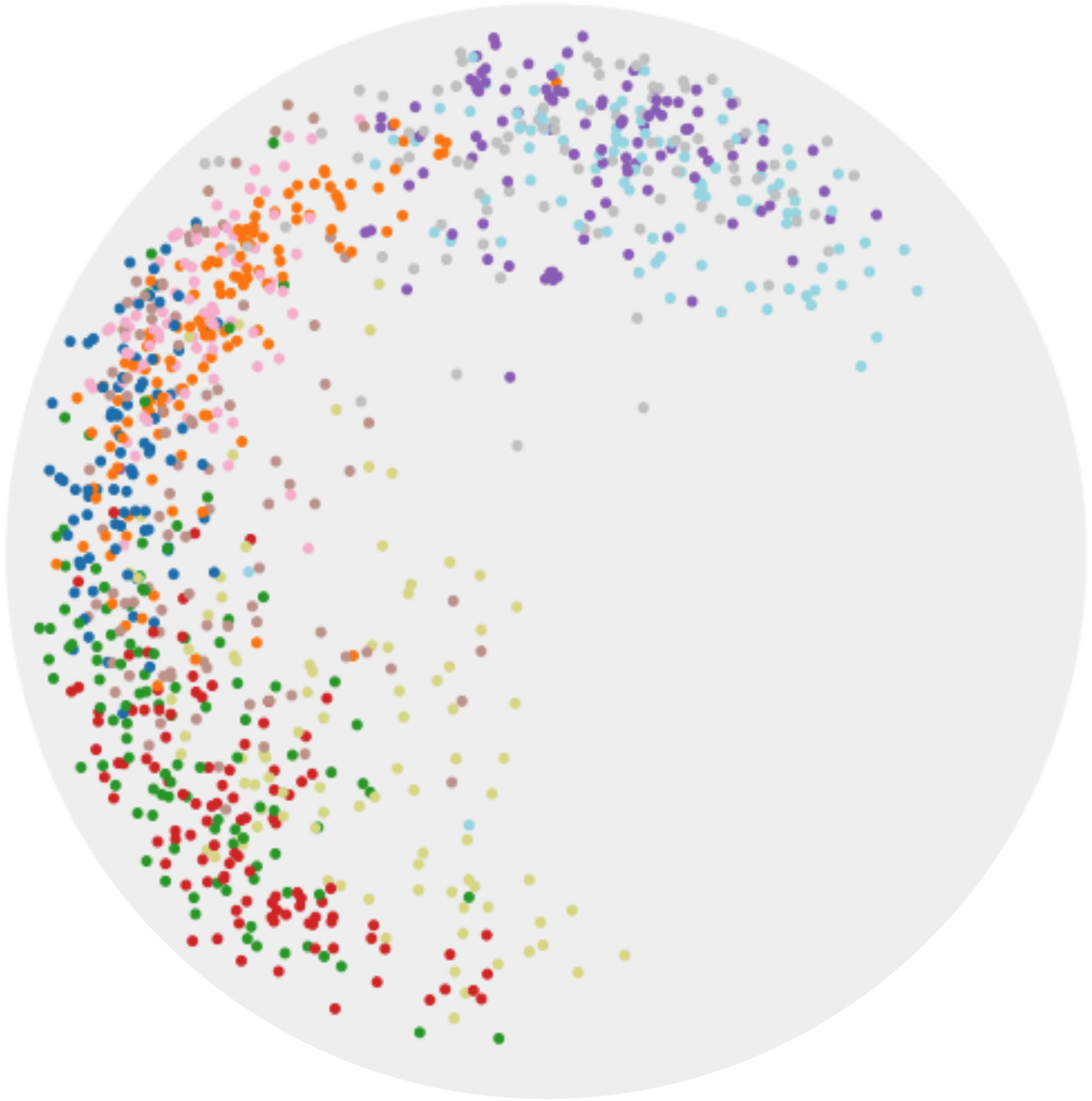} &
   \includegraphics[height=0.18\textheight]{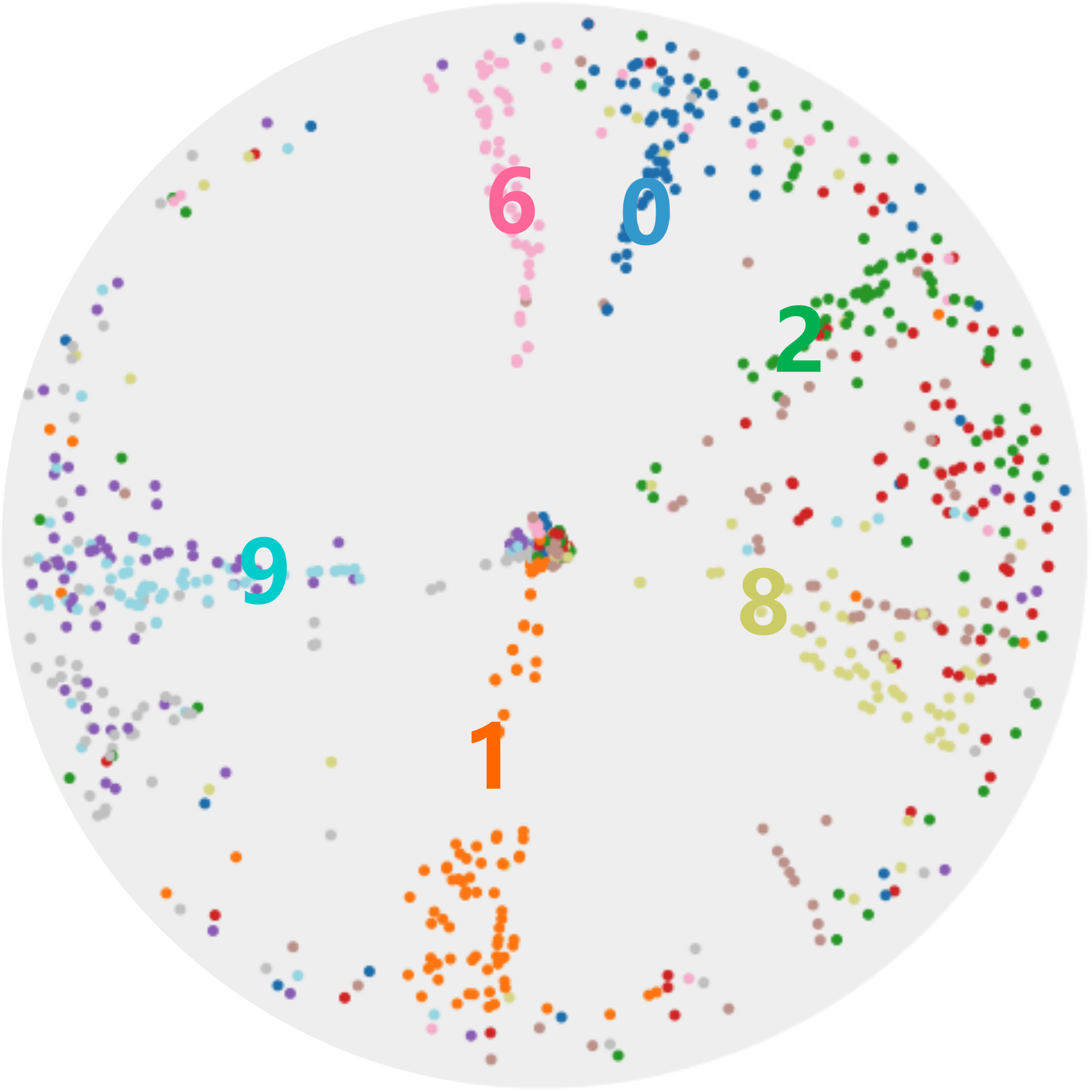} \\
   a) HoroPCA & b) CO-SNE
\end{tabular}
    \caption{CO-SNE produces better visualization of the high-dimensional latent space representations generated by the encoder of the Poincar\'e VAE than HoroPCA. a) The two-dimensional latent Poincar\'e embeddings generated from HoroPCA. b) The two-dimensional latent Poincar\'e embeddings generated from CO-SNE. CO-SNE can capture the hierarchical and clustering structures of the high-dimensional latent representations while HoroPCA cannot.}
    \label{fig:pvae}
\end{figure}

\section{Latent Space of Poincar\'e VAE}
\label{sec:v_pvae}
Variational autoencoders (VAEs) \cite{kingma2013auto} is a popular unsupervised learning method. Standard VAEs assume the latent space is Euclidean. \cite{mathieu2019continuous} extends VAEs by assuming the latent space is hyperbolic. Different from the standard VAEs, Poincar\'e VAEs can embed tree-like structures more efficiently. For using CO-SNE to visualize the latent space of Poincar\'e VAEs, we train a Poincar\'e VAE with a latent dimension of five on MNIST \cite{lecun1998mnist} following from \cite{mathieu2019continuous}. We further generate the latent space representations of 1000 randomly sampled images with the encoder.  

Figure \ref{fig:pvae} shows the visualizations produced by HoroPCA and CO-SNE. Clearly, CO-SNE produces much better visualization than HoroPCA. In particular, we can easily observe the hierarchical and clustering structures in the latent space which are totally distorted in the visualization produced by HoroPCA. Thus, CO-SNE can be used to understand the latent space of Poincar\'e VAEs and facilitate the development of better unsupervised hyperbolic learning methods.


\subsection{Discussion}
CO-SNE can be used to visualize hyperbolic datapoints in a two-dimensional hyperbolic space while maintaining the local similarities and hierarchical structure. CO-SNE still suffers from the same weaknesses as t-SNE. In particular, as a general method for dimensionality reduction, the curse of intrinsic dimensionality and the non-convexity, please see \cite{van2008visualizing} for more details. More ablations and results can be found in the Supplementary. In particular, we investigate the effect of hyperparameters for balancing the KL-divergence and the distance loss. 

\section{Summary}

We propose CO-SNE which can be used to visualize hyperbolic data in a two-dimensional hyperbolic space. We use the generalized versions of the normal and Cauchy distributions in hyperbolic space to compute the high-dimensional and the low-dimensional similarities, respectively. We apply CO-SNE to visualize synthetic hyperbolic data, biological data and embeddings learned by supervised and unsupervised representation learning methods. CO-SNE produces much better visualizations than several popular visualization methods.

{\bf Acknowledgements.} 
This work was supported, in part, by Berkeley Deep Drive and the National Science Foundation (NSF) under Grant No. 2131111. Any opinions, findings, and conclusions or recommendations expressed in this material are those of the authors and do not necessarily reflect the views of the NSF.

\section{Supplementary}

\subsection{Riemannian Geometry}
\label{sec:riemannian}
We give a brief overview of Riemannian geometry,  for more details please refer to \cite{carmo1992riemannian}. 
A Riemannian manifold $(\mathcal{M}, \mathfrak{g})$ is a real smooth manifold $\mathcal{M}$ with a Riemannian metric $\mathfrak{g}$. The Riemannian metric $\mathfrak{g}$ is a smoothly varying inner product which is defined on the tangent space $T_\mathbf{x}{\mathcal{M}}$ of $\mathcal{M}$. Given $\mathbf{x} \in \mathcal{M}$ and two vectors $ \mathbf{v}, \mathbf{w} \in T_\mathbf{x}{\mathcal{M}}$, we can use the Riemannian metric $\mathfrak{g}$ to compute the inner product  $\langle \mathbf{v}, \mathbf{w} \rangle_\mathbf{x}$ as $\mathfrak{g}(\mathbf{v}, \mathbf{w})$. The norm of $\mathbf{v} \in T_\mathbf{x}{\mathcal{M}}$ is defined as $\lVert \mathbf{v} \rVert_\mathbf{x} = \sqrt{ \langle \mathbf{v}, \mathbf{v} \rangle}_\mathbf{x}$. A geodesic generalizes the notion of straight line in the manifold which is defined as a curve $\gamma : [0, 1] \rightarrow \mathcal{M}$ of constant speed that is everywhere locally a distance minimizer. The exponential map and the inverse exponential map are defined as follows: given $\mathbf{x}, \mathbf{y} \in \mathcal{M}, \mathbf{v} \in T_\mathbf{x}{\mathcal{M}}$, and a geodesic $\gamma$ of length $\lVert \mathbf{v} \rVert$ such that $\gamma(0) = \mathbf{x}, \gamma(1) = \mathbf{y}, \gamma'(0) =  \mathbf{v} / \lVert  \mathbf{v} \rVert$, the exponential map $\textnormal{Exp}_\mathbf{x}: T_\mathbf{x}{\mathcal{M}} \rightarrow \mathcal{M}$ satisfies $\textnormal{Exp}_\mathbf{x}(\mathbf{v}) = \mathbf{y}$ and the inverse exponential map $\textnormal{Exp}^{-1}_\mathbf{x}: \mathcal{M} \rightarrow T_\mathbf{x}{\mathcal{M}}$ satisfies $\textnormal{Exp}^{-1}_\mathbf{x}(\mathbf{y}) = \mathbf{v}$.

\subsection{Hyperbolic Student's t-Distribution}
\label{sec:hs}
Recall the way to define the the Student's t-distribution which expresses the random variable $t$ as,
\begin{equation}
    t = \frac{u}{\sqrt{v/n}}
    \label{eq: t_variable_sup}
\end{equation}

where $u$ is a random variable sampled from a standard normal distribution and $v$ is a random variable sampled from a $\chi^2$-distribution of $n$ degrees of freedom. The $\chi^2$-distribution can also be derived from normal distribution. Let $u_1$, $u_2$, ..., $u_n$ be independent standard normal random variables, then the sum of the squares,

\begin{equation}
    v  = \sum_{i=1}^n u_i^2
\end{equation}
is a $\chi^2$-distribution with $n$ degrees of freedom. Thus, the probability density function of the $\chi^2$-distribution can be derived from the probability density function of the normal distribution which is,

\begin{equation}
    T_n(v) = \frac{1}{2^{n/2}\Gamma(n/2)} v^{(n-2)/2}e^{-v/2}  
\end{equation}

Using Equation \ref{eq: t_variable_sup}, we can further derive the probability density function of the Student's t-distribution,

\begin{equation}
    f_n(t) = \frac{1}{\sqrt{n}B(1/2, n/2)}(1+\frac{t^2}{n})^{-(n+1)/2}
\end{equation}
where $B$ is the Beta function. 

The probability density function of hyperbolic Cauchy distribution can be derived in a similar way using hyperbolic normal distribution.

\begin{figure*}
   \centering
   \setlength\tabcolsep{0.05pt}%
\begin{tabular}{c|ccccc}
& $\lambda_2$ = 0.0 & $\lambda_2$ = 0.01 & $\lambda_2$ = 0.05 & $\lambda_2$ = 0.1 & $\lambda_2$ = 0.2 \\ 
\toprule
$\lambda_1$ = 0 & \includegraphics[align=c,width=0.18\textwidth]{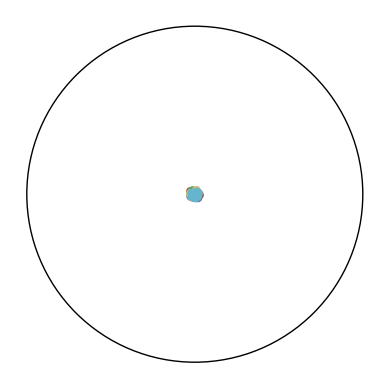}&
\includegraphics[align=c,width=0.18\textwidth]{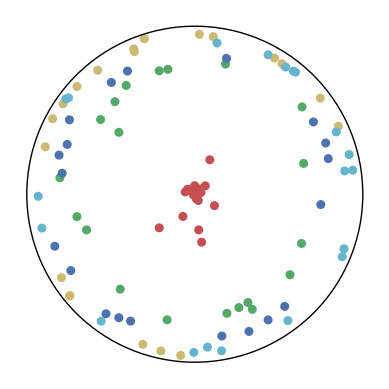}&
\includegraphics[align=c,width=0.18\textwidth]{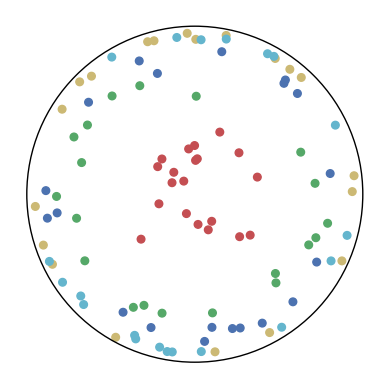}&
\includegraphics[align=c,width=0.18\textwidth]{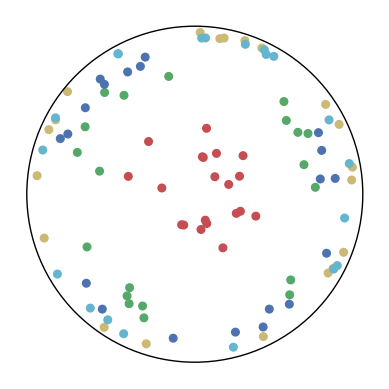} &
\includegraphics[align=c,width=0.18\textwidth]{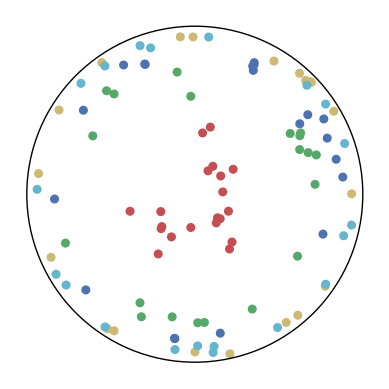}\\
\bottomrule

\centering  $\lambda_1$ = 5 &\includegraphics[align=c,width=0.18\textwidth]{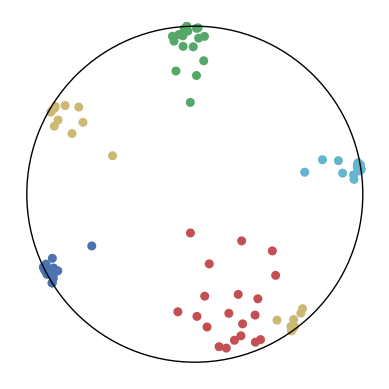}&
\includegraphics[align=c,width=0.18\textwidth]{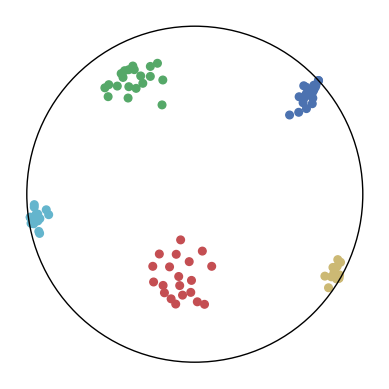}&
\includegraphics[align=c,width=0.18\textwidth]{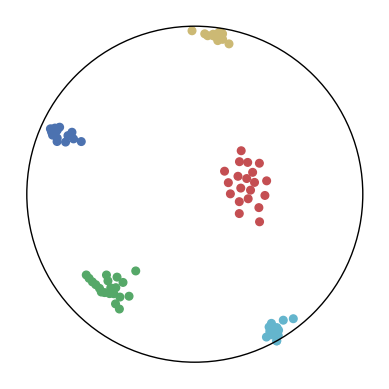}&
\includegraphics[align=c,width=0.18\textwidth]{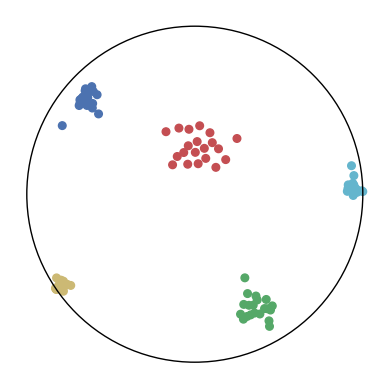} &
\includegraphics[align=c,width=0.18\textwidth]{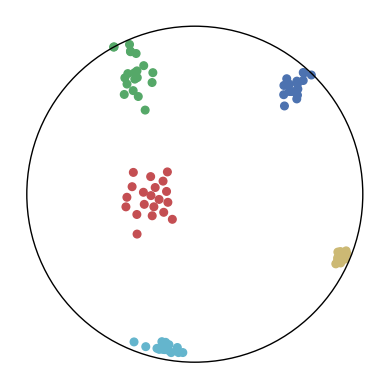}\\

\bottomrule
$\lambda_1$ = 10 &\includegraphics[align=c,width=0.18\textwidth]{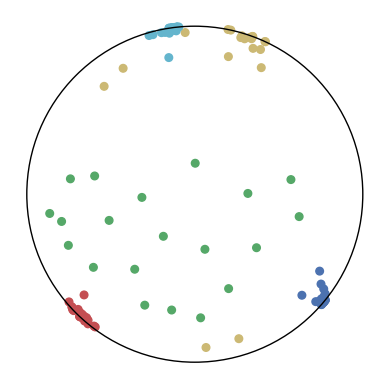}&
\includegraphics[align=c,width=0.18\textwidth]{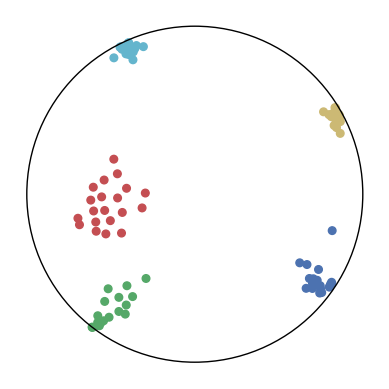}&
\includegraphics[align=c,width=0.18\textwidth]{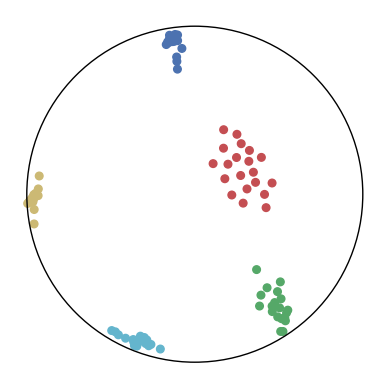}&
\includegraphics[align=c,width=0.18\textwidth]{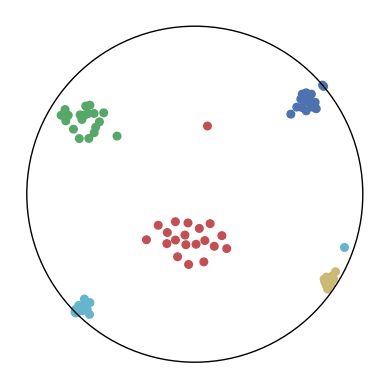} &
\includegraphics[align=c,width=0.18\textwidth]{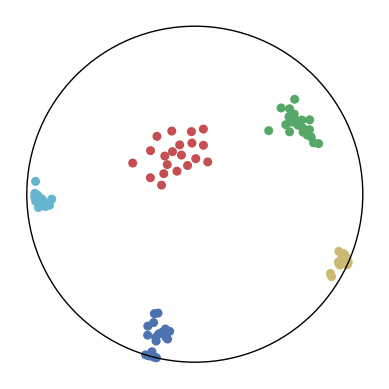}\\

\bottomrule
$\lambda_1 = 15$&\includegraphics[align=c,width=0.18\textwidth]{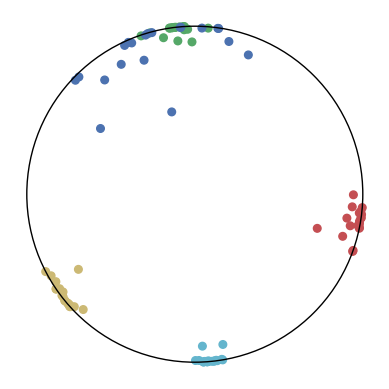}&
\includegraphics[align=c,width=0.18\textwidth]{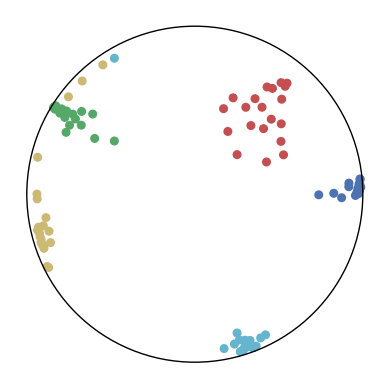}&
\includegraphics[align=c,width=0.18\textwidth]{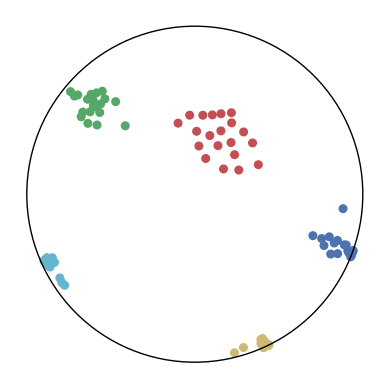}&
\includegraphics[align=c,width=0.18\textwidth]{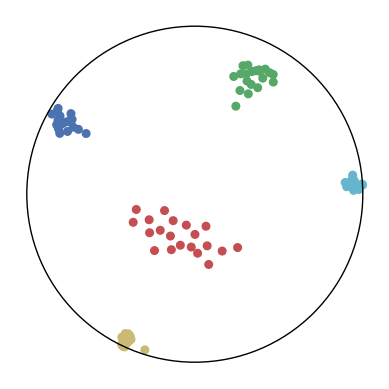} &
\includegraphics[align=c,width=0.18\textwidth]{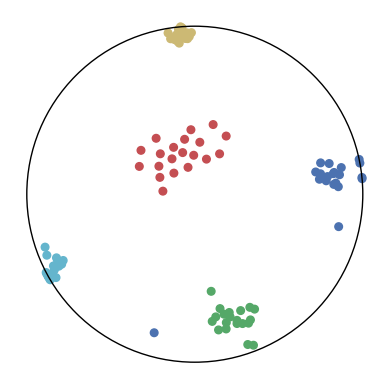}\\

\bottomrule
$\lambda_1$ = 20 &\includegraphics[align=c,width=0.18\textwidth]{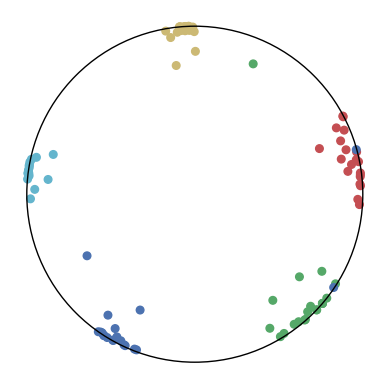}&
\includegraphics[align=c,width=0.18\textwidth]{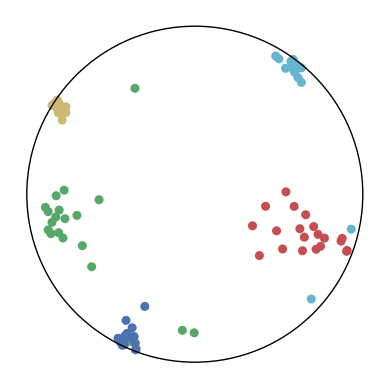}&
\includegraphics[align=c,width=0.18\textwidth]{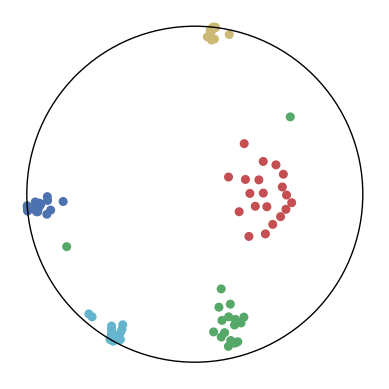}&
\includegraphics[align=c,width=0.18\textwidth]{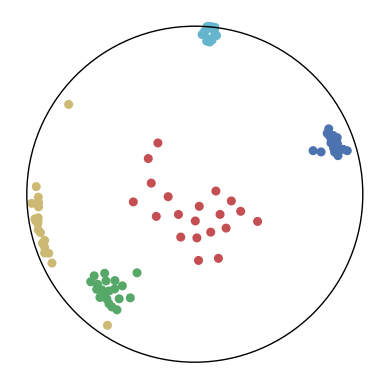} &
\includegraphics[align=c,width=0.18\textwidth]{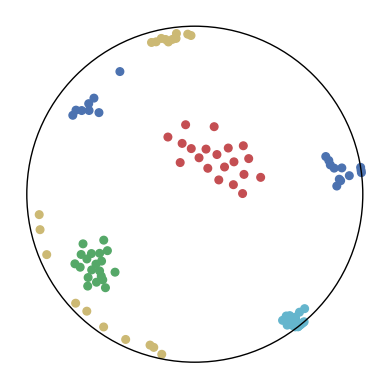}\\
\end{tabular}
    \caption{The effect of $\lambda_1$ and $\lambda_2$ on the quality of the visualization. We show the visualization results of CO-SNE on a mixture of five hyperbolic normal distributions in a five-dimensional hyperbolic space. We can observe that $\lambda_1$ is responsible for preserving the local similarity structure and $\lambda_2$ is responsible for preserving the global hierarchical structure. \emph{This shows that both the KL-divergence and the distance loss are important for producing good visualization}. CO-SNE is robust to wide choices of $\lambda_1$ and $\lambda_2$. Generally, $\lambda_1$ should be larger than $\lambda_2$ since the magnitude of the gradients of the KL-divergence is smaller. }
    \label{fig:ablation} 
\end{figure*}

\subsection{Hyperbolic Cauchy Distribution}
\label{sec:hc}
Similar to the Student's t-Distribution, the probability density function of Cauchy distribution can derived from the probability density function of the normal distribution. In particular, let $X$ and $Y$ be independent standard normal random variables, then $Z = \frac{X}{X + Y}$ is a Cauchy random variable. The probability density function of Cauchy distribution can be written as,

\begin{equation}
    f(x;x_0, \gamma) = \frac{1}{\pi \gamma [ 1 + (\frac{x-x_0}{\gamma})^2]}
\end{equation}
The probability density function of hyperbolic Cauchy distribution can be derived in a similar way using hyperbolic normal distribution.

The repulsion and attraction forces in CO-SNE depend on the term $p_{ij} - q_{ij}$ in Equation 12 of the main text. $p_{ij}$ is fixed during training which depends on the distribution of the high-dimensional datapoints. To create more repulsion forces between two close low-dimensional embeddings $y_i$ and $y_j$, we aim at increasing the probability that the point $y_i$ would select the point $y_j$ as its neighbor (i.e., $q_{ij}$). By using a small $\gamma$ in hyperbolic Cauchy distribution, the distance between $y_i$ and $y_j$ is scaled up. When the point $y_i$ is fixed, the probability of selecting $y_j$ as a neighbor (i.e., $q_{ij}$) is scaled up relatively to some point $y_k$ which is far away from $y_i$. As a result, $p_{ij} - q_{ij}$ can potentially become negative. This creates additional repulsion forces to push the low-dimensional points apart.

\subsection{The Effect of $\lambda_1$ and $\lambda_2$}
Recall the objective function of CO-SNE,
\label{sec:ablation}
\begin{equation}
    \mathcal{L} = \lambda_1 \mathcal{C} + \lambda_2 \mathcal{H}
\end{equation}
$\lambda_1$ and $\lambda_2$ are used to balance the KL-divergence $\mathcal{C}$ and the distance loss $\mathcal{H}$ and can be regarded as the learning rates. For ablation studies on the effect of $\lambda_1$ and $\lambda_2$, we reuse the settings in Section 4.1 of the main text.

We consider different settings of $\lambda_1$ and $\lambda_2$ to investigate the effect of the KL-divergence and the distance loss. The results are shown in Figure \ref{fig:ablation}. We have several observations from the results.

\begin{enumerate}
    \item In the first row, $\lambda_1 = 0.0$.  This means that only the distance loss is presented. We can observe that the low-dimensional embeddings can only approximate the magnitude of the high-dimensional datapoints but not the similarity structure.
    
    \item In the first column, $\lambda_2 = 0.0$. This means that only the KL-divergence is presented. We can observe that the low-dimensional embeddings can only preserve the similarity structure in the high-dimensional datapoints but not the hierarchical information.
    
    \item In other cases, we can observe that a larger $\lambda_2$ can better preserve the hierarchical structure in the high-dimensional datapoints but may distort the similarity structure. A larger $\lambda_1$ may also lead to a bad visualization of the similarity structure since the KL-divergence might diverge. Nevertheless, CO-SNE is robust to wide choices of $\lambda_1$ and $\lambda_2$. \emph{Both the  KL-divergence and the distance loss are important for producing good visualization}. 
\end{enumerate}

{\small
\bibliographystyle{ieee_fullname}
\bibliography{egbib}

\begin{thebibliography}{10}\itemsep=-1pt

\bibitem{alanis2016efficient}
Gregorio Alanis-Lobato, Pablo Mier, and Miguel~A Andrade-Navarro.
\newblock Efficient embedding of complex networks to hyperbolic space via their
  laplacian.
\newblock {\em Scientific reports}, 6(1):1--10, 2016.

\bibitem{arora2018analysis}
Sanjeev Arora, Wei Hu, and Pravesh~K Kothari.
\newblock An analysis of the t-sne algorithm for data visualization.
\newblock In {\em Conference On Learning Theory}, pages 1455--1462. PMLR, 2018.

\bibitem{bonnabel2013stochastic}
Silvere Bonnabel.
\newblock Stochastic gradient descent on riemannian manifolds.
\newblock {\em IEEE Transactions on Automatic Control}, 58(9):2217--2229, 2013.

\bibitem{carmo1992riemannian}
Manfredo Perdigao~do Carmo.
\newblock {\em Riemannian geometry}.
\newblock Birkh{\"a}user, 1992.

\bibitem{chami2021horopca}
Ines Chami, Albert Gu, Dat~P Nguyen, and Christopher R{\'e}.
\newblock Horopca: Hyperbolic dimensionality reduction via horospherical
  projections.
\newblock In {\em International Conference on Machine Learning}, pages
  1419--1429. PMLR, 2021.

\bibitem{cho2019large}
Hyunghoon Cho, Benjamin DeMeo, Jian Peng, and Bonnie Berger.
\newblock Large-margin classification in hyperbolic space.
\newblock In {\em The 22nd International Conference on Artificial Intelligence
  and Statistics}, pages 1832--1840. PMLR, 2019.

\bibitem{ganea2018hyperbolic}
Octavian-Eugen Ganea, Gary B{\'e}cigneul, and Thomas Hofmann.
\newblock Hyperbolic neural networks.
\newblock {\em arXiv preprint arXiv:1805.09112}, 2018.

\bibitem{gulcehre2018hyperbolic}
Caglar Gulcehre, Misha Denil, Mateusz Malinowski, Ali Razavi, Razvan Pascanu,
  Karl~Moritz Hermann, Peter Battaglia, Victor Bapst, David Raposo, Adam
  Santoro, et~al.
\newblock Hyperbolic attention networks.
\newblock {\em arXiv preprint arXiv:1805.09786}, 2018.

\bibitem{guo2021free}
Yunhui Guo, Xudong Wang, Yubei Chen, and Stella~X Yu.
\newblock Free hyperbolic neural networks with limited radii.
\newblock {\em arXiv preprint arXiv:2107.11472}, 2021.

\bibitem{gupte2011finding}
Mangesh Gupte, Pravin Shankar, Jing Li, Shanmugauelayut Muthukrishnan, and
  Liviu Iftode.
\newblock Finding hierarchy in directed online social networks.
\newblock In {\em Proceedings of the 20th international conference on World
  wide web}, pages 557--566, 2011.

\bibitem{hsu2020learning}
Joy Hsu, Jeffrey Gu, Gong-Her Wu, Wah Chiu, and Serena Yeung.
\newblock Learning hyperbolic representations for unsupervised 3d segmentation.
\newblock {\em arXiv preprint arXiv:2012.01644}, 2020.

\bibitem{jolliffe2016principal}
Ian~T Jolliffe and Jorge Cadima.
\newblock Principal component analysis: a review and recent developments.
\newblock {\em Philosophical Transactions of the Royal Society A: Mathematical,
  Physical and Engineering Sciences}, 374(2065):20150202, 2016.

\bibitem{khrulkov2020hyperbolic}
Valentin Khrulkov, Leyla Mirvakhabova, Evgeniya Ustinova, Ivan Oseledets, and
  Victor Lempitsky.
\newblock Hyperbolic image embeddings.
\newblock In {\em Proceedings of the IEEE/CVF Conference on Computer Vision and
  Pattern Recognition}, pages 6418--6428, 2020.

\bibitem{kingma2013auto}
Diederik~P Kingma and Max Welling.
\newblock Auto-encoding variational bayes.
\newblock {\em arXiv preprint arXiv:1312.6114}, 2013.

\bibitem{klimovskaia2020poincare}
Anna Klimovskaia, David Lopez-Paz, L{\'e}on Bottou, and Maximilian Nickel.
\newblock Poincar{\'e} maps for analyzing complex hierarchies in single-cell
  data.
\newblock {\em Nature communications}, 11(1):1--9, 2020.

\bibitem{lecun1998mnist}
Yann LeCun.
\newblock The mnist database of handwritten digits.
\newblock {\em http://yann. lecun. com/exdb/mnist/}.

\bibitem{liu2019hyperbolic}
Qi Liu, Maximilian Nickel, and Douwe Kiela.
\newblock Hyperbolic graph neural networks.
\newblock {\em arXiv preprint arXiv:1910.12892}, 2019.

\bibitem{mathieu2019continuous}
Emile Mathieu, Charline~Le Lan, Chris~J Maddison, Ryota Tomioka, and Yee~Whye
  Teh.
\newblock Continuous hierarchical representations with poincar$\backslash$'e
  variational auto-encoders.
\newblock {\em arXiv preprint arXiv:1901.06033}, 2019.

\bibitem{mcinnes2018umap}
Leland McInnes, John Healy, and James Melville.
\newblock Umap: Uniform manifold approximation and projection for dimension
  reduction.
\newblock {\em arXiv preprint arXiv:1802.03426}, 2018.

\bibitem{mcinnes2018umap-software}
Leland McInnes, John Healy, Nathaniel Saul, and Lukas Grossberger.
\newblock Umap: Uniform manifold approximation and projection.
\newblock {\em The Journal of Open Source Software}, 3(29):861, 2018.

\bibitem{miller1995wordnet}
George~A Miller.
\newblock Wordnet: a lexical database for english.
\newblock {\em Communications of the ACM}, 38(11):39--41, 1995.

\bibitem{nagano2019wrapped}
Yoshihiro Nagano, Shoichiro Yamaguchi, Yasuhiro Fujita, and Masanori Koyama.
\newblock A wrapped normal distribution on hyperbolic space for gradient-based
  learning.
\newblock In {\em International Conference on Machine Learning}, pages
  4693--4702. PMLR, 2019.

\bibitem{nickel2017poincar}
Maximilian Nickel and Douwe Kiela.
\newblock Poincar$\backslash$'e embeddings for learning hierarchical
  representations.
\newblock {\em arXiv preprint arXiv:1705.08039}, 2017.

\bibitem{olsson2016singlecell}
Andre Olsson, Meenakshi Venkatasubramanian, Viren~K Chaudhri, Bruce~J Aronow,
  Nathan Salomonis, Harinder~1 Singh, and H.~Leighton Grimes.
\newblock Single-cell analysis of mixed-lineage states leading to a binary cell
  fate choice.
\newblock {\em Nature}, 2016.

\bibitem{scikit-learn}
F. Pedregosa, G. Varoquaux, A. Gramfort, V. Michel, B. Thirion, O. Grisel, M.
  Blondel, P. Prettenhofer, R. Weiss, V. Dubourg, J. Vanderplas, A. Passos, D.
  Cournapeau, M. Brucher, M. Perrot, and E. Duchesnay.
\newblock Scikit-learn: Machine learning in {P}ython.
\newblock {\em Journal of Machine Learning Research}, 12:2825--2830, 2011.

\bibitem{pennec2006intrinsic}
Xavier Pennec.
\newblock Intrinsic statistics on riemannian manifolds: Basic tools for
  geometric measurements.
\newblock {\em Journal of Mathematical Imaging and Vision}, 25(1):127--154,
  2006.

\bibitem{roweis2000nonlinear}
Sam~T Roweis and Lawrence~K Saul.
\newblock Nonlinear dimensionality reduction by locally linear embedding.
\newblock {\em science}, 290(5500):2323--2326, 2000.

\bibitem{shimizu2020hyperbolic}
Ryohei Shimizu, Yusuke Mukuta, and Tatsuya Harada.
\newblock Hyperbolic neural networks++.
\newblock {\em arXiv preprint arXiv:2006.08210}, 2020.

\bibitem{tenenbaum2000global}
Joshua~B Tenenbaum, Vin De~Silva, and John~C Langford.
\newblock A global geometric framework for nonlinear dimensionality reduction.
\newblock {\em science}, 290(5500):2319--2323, 2000.

\bibitem{ungar2008gyrovector}
Abraham~Albert Ungar.
\newblock A gyrovector space approach to hyperbolic geometry.
\newblock {\em Synthesis Lectures on Mathematics and Statistics}, 1(1):1--194,
  2008.

\bibitem{van2008visualizing}
Laurens Van~der Maaten and Geoffrey Hinton.
\newblock Visualizing data using t-sne.
\newblock {\em Journal of machine learning research}, 9(11), 2008.

\bibitem{weber2020robust}
Melanie Weber, Manzil Zaheer, Ankit~Singh Rawat, Aditya Menon, and Sanjiv
  Kumar.
\newblock Robust large-margin learning in hyperbolic space.
\newblock {\em arXiv preprint arXiv:2004.05465}, 2020.

\bibitem{weng2021unsupervised}
Zhenzhen Weng, Mehmet~Giray Ogut, Shai Limonchik, and Serena Yeung.
\newblock Unsupervised discovery of the long-tail in instance segmentation
  using hierarchical self-supervision.
\newblock In {\em Proceedings of the IEEE/CVF Conference on Computer Vision and
  Pattern Recognition}, pages 2603--2612, 2021.

\bibitem{zhou2020graph}
Jie Zhou, Ganqu Cui, Shengding Hu, Zhengyan Zhang, Cheng Yang, Zhiyuan Liu,
  Lifeng Wang, Changcheng Li, and Maosong Sun.
\newblock Graph neural networks: A review of methods and applications.
\newblock {\em AI Open}, 1:57--81, 2020.

\end{thebibliography}
}

\end{document}